\documentclass{article}

\usepackage[nonatbib,final]{neurips_2020}

\usepackage[utf8]{inputenc} % allow utf-8 input
\usepackage[T1]{fontenc}    % use 8-bit T1 fonts
\usepackage{hyperref}       % hyperlinks
\usepackage{url}            % simple URL typesetting
\usepackage{booktabs}       % professional-quality tables
\usepackage{amsfonts}       % blackboard math symbols
\usepackage{nicefrac}       % compact symbols for 1/2, etc.
\usepackage{microtype}      % microtypography
\usepackage{color}
\usepackage{graphicx}
\usepackage{subfigure}
\usepackage{multirow}
\usepackage{cite}
\usepackage{amsmath}

\usepackage{algorithm}
\usepackage{algorithmicx}
\usepackage{algpseudocode}

\usepackage[english]{babel}
\usepackage{amsthm}
\theoremstyle{definition}
\newtheorem{definition}{Definition}[section]

\MakeRobust{\Call}
\floatname{algorithm}{Algorithm}

\title{The Limit of the Batch Size}

\author{%
  Yang You$^1$, Yuhui Wang$^1$, Huan Zhang$^2$, Zhao Zhang$^3$, James Demmel$^1$, Cho-Jui Hsieh$^2$\\
  UC Berkeley$^1$, UCLA$^2$, TACC$^3$ \\
  \texttt{\{youyang, demmel\}@cs.berkeley.edu}, yuhui-w@berkeley.edu, \\huanzhang@ucla.edu, zzhang@tacc.utexas.edu, chohsieh@cs.ucla.edu\\
}

\begin{document}

\maketitle

\begin{abstract}

Large-batch training is an efficient approach for current distributed deep learning systems.
It has enabled researchers to reduce the ImageNet/ResNet-50 training time from 29 hours to around 1 minute.
In this paper, we focus on studying the limit of the batch size.
We think it may provide a guidance to AI supercomputer and algorithm designers.
We provide detailed numerical optimization instructions for step-by-step comparison.
Moreover, it is important to understand the generalization and optimization performance of huge batch training.
Hoffer et al. \cite{hoffer2017train} introduced the ``ultra-slow diffusion'' theory to large-batch training.
However, our experiments show contradictory results with the conclusion of \cite{hoffer2017train}.
We provide comprehensive experimental results and detailed analysis to study the limitations of batch size scaling and ``ultra-slow diffusion'' theory.
For the first time we scale the batch size on ImageNet to at least a magnitude larger than all previous work, and provide detailed studies on the performance of many state-of-the-art optimization schemes under this setting. We propose an optimization recipe that is able to improve the top-1 test accuracy by 18\% compared to the baseline.

%We report a state-of-the-art accuracy for huge-batch and full-batch training.
%These exciting results have encouraged industry to design and implement several AI supercomputers like Google TPU Pod, NVIDIA SuperPOD, and Huawei Atlas 900 AI Cluster.
%Based on the experimental results and analysis, our study leads to three conjectures: (1) For a realistic dataset/model training (e.g. ImageNet/ResNet-50), there exists a huge-batch region; (2) There is boundary between large-batch and huge-batch. Above this boundary, the learning process becomes totally different; (3) For a realistic dataset/model with full-batch size, we can not reach the target accuracy in polynomial time by current optimization techniques.

\end{abstract}

\section{Introduction}

Large-batch optimization is becoming an important research topic as it is an efficient approach for current distributed deep learning systems.
It has enabled researchers to reduce the ImageNet/ResNet-50 training from 29 hours to around 1 minute (Table \ref{table:imagenet_resnet50}).
Researchers can also reduce the BERT pre-training time from 3 days to 76 minutes \cite{you2019large}, 67 minutes \cite{shoeybi2019megatron}, 47 minutes \cite{nvidia2019bert} and 44 minutes \cite{rajbhandari2019zero}.
The speedup comes from the fact that researchers can scale the training to a larger batch size, and so use larger scale, parallelism, without losing accuracy in a fixed number of epochs.
For example, researchers scaled the batch size of ImageNet training from 256 \cite{he2016deep} to 1K \cite{krizhevsky2014one}, 5K \cite{li2017scaling}, 8K \cite{goyal2017accurate}, 16K \cite{you2017imagenet}, 32K \cite{ying2018image}, and 64K \cite{kumar2019scale}.
On the other hand, the hardware designers are also interested in large-batch training, as the chip makers fail to continue improving the single processor's clock frequency due to power consumption issue \cite{asanovic2009view}.
Thus, the hardware vendors have to use more processors to increase the raw floating point computing performance, which requires high-parallelism for future algorithms.
Many successful use cases of large-batch training have encouraged industry to design and implement several AI supercomputers like Google TPU Pod, NVIDIA SuperPOD, and Huawei Atlas 900 AI Cluster to take advantage of the extremely high parallelism provided by large-batch algorithms.

\begin{table}[ht]
\renewcommand{\arraystretch}{1.3}
\caption{\footnotesize ImageNet/ResNet-50 Training Speed Records.}
\centering
\tiny
\begin{tabular}{|c|c|c|c|c|c|}
\hline
Teams & Date & Accuracy & Time & {Optimizer}\\
\hline
\hline
He et al. \cite{he2016deep} & 12/10/2015 & 75.3\% & 29h & Momentum (Rumelhart et al. \cite{rumelhart1986learning})\\
\hline
Goyal et al. \cite{goyal2017accurate} & 06/08/2017 & 76.3\% & 65m & Momentum (Rumelhart et al. \cite{rumelhart1986learning})\\
\hline
You et al. \cite{you2017imagenet} & 11/02/2017 & 75.3\% & 48m & LARS (You et al. \cite{you2017scaling})\\
\hline
You et al. \cite{you2017imagenet} & 11/07/2017 & 75.3\% & 31m & LARS (You et al. \cite{you2017scaling}) \\
\hline
Akiba et al. \cite{akiba2017extremely} & 11/12/2017 & 74.9\% & 15m & RMSprop (Hinton \cite{tieleman2012lecture}) \\
\hline
You et al. \cite{you2017imagenet} & 12/07/2017 & 74.9\% & 14m & LARS (You et al. \cite{you2017scaling}) \\
\hline
Jia et al. \cite{jia2018highly} & 07/30/2018 & 75.8\% & 6.6m & LARS (You et al. \cite{you2017scaling}) \\
\hline
Mikami et al. \cite{mikami2018imagenet} & 11/14/2018 & 75.0\% & 3.7m & LARS (You et al. \cite{you2017scaling}) \\
\hline
Ying et al. \cite{ying2018image} & 11/16/2018 & 76.3\% & 2.2m & LARS (You et al. \cite{you2017scaling}) \\
\hline
Yamazaki et al. \cite{yamazaki2019yet} & 03/29/2019 & 75.1\% & 1.25m & LARS (You et al. \cite{you2017scaling}) \\
\hline
Kumar et al. \cite{kumar2019scale} & 10/02/2019 & 75.9\% & 67.1s & LARS (You et al. \cite{you2017scaling}) \\
\hline
\end{tabular}
\label{table:imagenet_resnet50}
\end{table}

In this paper, we focus on studying the limit of the batch size and hope to provide guidance to AI supercomputer and algorithm designers. To the best of our knowledge, we have not found any previous works studying a batch size over 128K.
We pick some realistic applications like ImageNet/ResNet-50, and provide detailed numerical optimization instructions for step-by-step accuracy comparison.
Moreover, it is important to understand the generalization and optimization difficulties encountered when batch size is huge.
There are a series of works \cite{hoffer2017train} \cite{keskar2016large} \cite{smith2017don} on this topic.
Specifically, Hoffer et al. \cite{hoffer2017train} introduced ``ultra-slow diffusion'' theory to large-batch training.
They pointed out that there is an ``ultra-slow'' logarithmic increase in the distance of the weights from their initialization.
Their paper indicated that, under a series of optimization techniques, the generalization performance can be controlled by the number of iterations.
However, our experiments show contradictory results once the batch size is increased beyond a certain boundary, which we refer to as the ``huge-batch'' regime.
We provide comprehensive experimental results and detailed analysis to study the limitations of batch size scaling and the ``ultra-slow diffusion'' theory.

Concretely, we scale the batch size of Imageanet/ResNet-50 to 819K and 1.28 million, which is an order of magnitude larger than any previous works. 
This is the first work to report an accuracy for huge/full-batch ImageNet/ResNet-50 training.
We also scale the batch size to the full-dataset for MNIST, CIFAR-10, and ImageNet\textcolor{blue}. The contribution of our paper can be summarized as follows:
\begin{itemize}
\item For the first time we scale the batch size on ImageNet to at least a magnitude larger that all previous works, and provide detailed studies on the performance of many state-of-the-art optimization schemes under this setting. We propose an optimization recipe that is able to improve the top-1 test accuracy by 18\% compared to the baseline.
\item We identify a \emph{``huge-batch'' regime} for realistic optimization tasks (e.g. ImageNet/ResNet-50) , where the optimization process becomes intrinsically harder and we cannot reach the target test accuracy in polynomial time by applying any of current optimization techniques. The ``ultra-slow diffusion'' theory does not hold any more in this regime.
\item Our results help system researchers and designers to understand the limit of parallelism in realistic machine learning tasks and design future machine learning hardware accordingly.
\end{itemize}

\iffalse
Based on the experimental results and analysis, our study leads to three conjectures:

\begin{itemize}
    \item For a realistic optimization task (e.g. ImageNet/ResNet-50), there exists a huge-batch range.
    \item There is boundary between large-batch and huge-batch. Above this boundary, the learning process becomes totally different.
    \item For a realistic dataset/model with full-batch size, we can not reach the target accuracy in polynomial time by current optimization techniques.
\end{itemize}

\begin{itemize}
    \item There is boundary between large-batch and huge-batch. Above this boundary, the learning process becomes a totally different optimization problem.
    \item There is an inherent generalization gap in huge-batch training and full-batch training for a realistic application like ImageNet/ResNet-50.
    \item For a realistic application with full batch size, we can not reach the target accuracy in polynomial time by current optimization techniques.
\end{itemize}

Definitions:
\begin{itemize}
    \item Large-batch: batch size between 2k and 64k. currently large-batch training is very stable and can achieve a good speedup
    \item Huge-batch: batch size over 64k
    \item Full-batch: batch size is equal to the total number of samples
\end{itemize}
\fi

\section{Background and Related Work\label{sec:related_work}}

For any vector $x_t \in \mathbb{R}^d$, either $x_{t,j}$ or $[x_t]_{j}$ are used to denote its $j^{\text{th}}$ coordinate where $j \in [d]$.  
%Let $\mathbb{I}$ be the $d \times d$ identity matrix, and let $\mathbb{I} = [\mathbb{I}_1, \mathbb{I}_2,...,\mathbb{I}_h]$ be its decomposition into column submatrices $\mathbb{I}_i = d \times d_h$. 
%For $x \in \mathbb{R}^d$, let $x^{(i)}$ be the block of variables corresponding to the columns of $I_i$ i.e., $x^{(i)} = \mathbb{I}_i^\top x \in \mathbb{R}^{d_i}$  for $i = \{1, 2, \cdots ,h\}$. 
For any function $f:\mathbb{R}^d \rightarrow \mathbb{R}$, we use $\nabla_i f(x)$ to denote the gradient with respect to $x^{(i)}$. 
%For any vectors $u, v \in \mathbb{R}^d$, we use $u^2$ and $u/v$ to denote elementwise square and division operators respectively.
We use $\|.\|$ to denote $l_2$-norm of a vector.
%We start our discussion by formally stating the problem setup.  
We study nonconvex stochastic optimization problems of the form
\begin{align}
\label{eq:1}
\min_{x \in \mathbb{R}^d} f(x) := \mathbb{E}_{s \sim \mathbb{P}}[\ell(x, s)] + \frac{\lambda}{2} \|x\|^2,
\end{align}
where $\ell$ is a smooth (possibly nonconvex) function and $\mathbb{P}$ is a probability distribution on the domain $\mathcal{S} \subset \mathbb{R}^k$. 
Here, $x$ corresponds to model parameters, $\ell$ is the loss function and $\mathbb{P}$ is an unknown data distribution.
\iffalse
We assume function $\ell(x)$ is $L$-\emph{smooth}, i.e.,  there exists a constant $L$ such that
\begin{equation}
\label{eq:l-const}
  \|\nabla_i \ell(x, s)-\nabla_i \ell(y, s)\| \le L\|x - y\|,\quad\forall\ x \in \mathbb{R}^d, y \in \mathbb{R}^d \text{ and  } s \in \mathcal{S},
\end{equation}
%for all $i \in [h]$. We use $L = (L_1, \cdots, L_h)^\top$ to denote the $h$-dimensional vector of Lipschitz constants. 
%We use $L_\infty$ and $L_{avg}$ to denote $\max_i L_i$ and $\sum_i \tfrac{L_i}{h}$ respectively.  
%We assume the following bound on the variance in stochastic gradients: $\mathbb{E}\|\nabla_i \ell(x, s) - \nabla_i f(x)\|^2 \leq \sigma_i^2$ for all $x \in \mathbb{R}^d$ and $i \in [h]$.    
We assume $\mathbb{E}\|[\nabla \ell(x, s)]_i - [\nabla f(x)]_i\|^2 \leq \tilde{\sigma}_i^2$ for all $x \in \mathbb{R}^d$ and $i \in [d]$.  
%We use $\sigma = (\sigma_1, \cdots, \sigma_h)^\top$ and $\tilde{\sigma} = (\tilde{\sigma}_1, \cdots, \tilde{\sigma}_d)^\top$ to denote the vectors of standard deviations of stochastic gradient per layer and per dimension respectively.  
We use $\tilde{\sigma} = (\tilde{\sigma}_1, \cdots, \tilde{\sigma}_d)^\top$ to denote the vectors of standard deviations of stochastic gradient per dimension.
Finally, we assume that the gradients are bounded i.e., $|[\nabla \ell(x,s)]_j| \leq G$ for all $j \in [d]$, $x \in \mathbb{R}^d$ and $s \in \mathcal{S}$. 
Note that such assumptions are typical in the analysis of stochastic first-order methods (cf. \cite{ghadimi2013stochastic,Reddi18}). 
\fi
Stochastic gradient descent (SGD) is a method for solving problem in Equation~(\ref{eq:1}). 
The update at the $t^{\text{th}}$ iteration of SGD is of the following form:
\begin{equation}
\tag{SGD}
x_{t+1} = x_{t} - \eta_t  \frac{1}{|\mathcal{S}_t|} \sum_{s_t \in \mathcal{S}_t} \nabla \ell(x_t, s_t) + \lambda x_t,
\end{equation}
where $S_t$ is set of $B$ random samples drawn from the  distribution $\mathbb{P}$. 
There are some SGD variants (e.g. \cite{duchi2011adaptive} \cite{kingma2014adam} \cite{tieleman2012lecture}) proposed in recent years.
%We define following terms in this paper.
We assume the batch size of our baseline is $B_0$, and our baseline can achieve a testing accuracy of $A$ with a validation loss of $\xi$ by a learning rate of $\eta$ in $\rho$ epochs.
We define define three optimization regimes with batch sizes $B_L$, $B_H$ and $B_F$:
\theoremstyle{definition}
\begin{definition}{\bf \textit{Large Batch}:}

If we use a batch size $B$ ($B < B_L, B \in Z^+$) and a learning rate of $\frac{B\eta}{B_0}$, then we can get a testing accuracy of $99.5\%A$ (or higher) and a validation loss of $1.2\xi$ (or lower) in $\rho$ epochs.
We define a batch size $B$ is large batch if $B \geq B_L, B \in Z^+$ (examples in Table \ref{table:baseline_application}).
%Straightforward optimization techniques can not get the target accuracy in the fixed number of epochs.
\end{definition}
\theoremstyle{definition}
\begin{definition}{\bf \textit{Huge Batch}:}
%We assume the batch size of our baseline is $B_0$, and our baseline can achieve a testing accuracy of $A$ with a training loss of $\xi$ by a learning rate of $\eta$ in $\rho$ epochs.
%We define a positive integer $B_H \in Z^+$.
If we use a batch size $B$ ($B > B_H, B \in Z^+$) with current available optimization techniques, then we can not get a testing accuracy of $99.5\%A$ in $\rho$ epochs.
We define a batch size $B$ is huge batch if $B > B_H, B \in Z^+$ (examples in Table \ref{table:baseline_application}).
\end{definition}
\theoremstyle{definition}
\begin{definition}{\bf \textit{Full Batch}:}
We define a positive integer $B_F \in Z^+$, which is equal to the total number of available training samples at runtime (examples in Table \ref{table:baseline_application}).
\end{definition}
%For very large batch settings, the following is a well-known result for $\sgd$.
Keskar et al. \cite{keskar2016large} reported that traditional first-order optimization techniques fail to scale up the batch size to a very large number.
Large-batch training will hurt the testing accuracy.
They indicate there is an inherent generalization gap in large-batch optimization {\bf within fixed number of epochs}.
Hoffer et al. \cite{hoffer2017train} suggested training longer can close the generalization gap.
After that, researchers have developed a series of optimization techniques and scaled the batch size to 64K without losing accuracy for various applications {\bf within fixed number of epochs} (Table \ref{table:imagenet_resnet50}). 
%\textcolor{blue}{include all large-batch training citations here} \cite{you2019large}
%On the other hand, Shallue et al. studied the relationship between batch size and number of steps required to reach a target accuracy. 
%A series of recent works are summarized in Table \ref{table:imagenet_resnet50}.
There are also other related literatures \cite{anil2019memory} \cite{ginsburg2019stochastic} \cite{gupta2020stochastic} \cite{golmant2018computational} \cite{krizhevsky2014one} \cite{lin2018don} \cite{liu2019variance} \cite{ma2019inefficiency} \cite{osawa2019large} \cite{shallue2018measuring} \cite{smith2017don}.
Inspired by these works, we focus on a different direction.
As far as we know, there is no previous work studying huge-batch optimization (e.g. over 128K).
Our investigation includes: (1) pushing the batch size to the limit and maximizing the accuracy in fixed number of epochs; (2) studying the optimization problems of huge-batch and full-batch; (3) studying the relationship between generalization and optimization.
%\begin{itemize}
%    \item Pushing the batch size to the limit and maximizing the accuracy in fixed number of epochs.
%    \item Studying the optimization problems of large-batch, huge-batch, and full-batch.
%    \item Studying the relationship between generalization and optimization.
%\end{itemize}

%\begin{Conjecture}{The limit of batch size}
%There is an upper bound of batch size in any deep neural networks training.
%We assume the batch size of our baseline is $B_0$, and our baseline can achieve a testing accuracy of $A$ with a training loss of $\xi$ by a learning rate of $\eta$ in $\rho$ epochs.
%We define a positive integer $B_H \in Z^+$.
%If we use a batch size $B$ ($B \geq B_H$ $\&\&$ $B \in Z^+$) with current available optimization techniques, then we can not get a testing accuracy of $[99.5\%A, 100.5\%A]$ in $\rho$ epochs.
%\end{Conjecture}

\iffalse
\begin{table}[ht]
\renewcommand{\arraystretch}{1.3}
\caption{\footnotesize The relationship between batch size and datasets.}
\centering
\tiny
\begin{tabular}{|c|c|c|c|c|c|}
\hline
Dataset & Baseline Batch & Large Batch & Huge Batch & Full Batch\\
\hline
\hline
ImageNet & 256 & [4K, 64K] & (64K, 1.28M) & 1.28M\\
\hline
MNIST & 256 & [2K, 8K] & (8k, 60k) & 60K \\
\hline
CIFAR-10 & 128 & [*K, *K] & (*k, 50k) & 50K \\
\hline
\end{tabular}
\label{table:dataset_batch}
\end{table}
\fi

\section{Optimization}
In this section, we use a series of techniques to maximize the accuracy in huge/full-batch training.
The information about our applications are in Table \ref{table:baseline_application}.
%In terms of ImageNet, we use v3-256 (128 chips), v3-1024 TPUs (512 chips) and v3-2048 TPUs (1024 chips) for 32K-batch, 819200-batch and full-batch ResNet-50 training, respectively.
%Other experiments are conducted on a system with 16 NVIDIA V100 GPUs.
Due to lack of space, some experimental results are in the Appendix.
The first step of our study is to identify the ``huge-batch'' regime, where we conduct experiments with extremely large batch sizes. We detail the process of finding the ``huge-batch'' numbers in Table \ref{table:baseline_application} and more results on how accuracy drops with batch size in the Appendix.

\begin{table}[ht]
\renewcommand{\arraystretch}{1.3}
\caption{\footnotesize The baseline information of our applications (e.g. For ImageNet, $B_L$=2K, $B_H$=64K, $B_F$=1.28M).}
\centering
\tiny
\begin{tabular}{|c|c|c|c|c|c|c|c|c|c|c|c|}
\hline
Dataset & Model & Implementation & Baseline Batch & Epochs & Top-1 Accuracy & Large Batch & Huge Batch & Full Batch\\
\hline
\hline
ImageNet & ResNet-50 v1.5 \cite{he2016deep} & \cite{resnet2019mlperf} & 256 & 90 & 75.9\% & [2K, 64K] & (64K, 1.28M) & 1.28M\\
\hline
MNIST & LeNet \cite{lecun2015lenet} & \cite{lenet2019google} & 256 & 30 & 99.2\% & [2K, 8K] & (8k, 60k) & 60K  \\
\hline
CIFAR-10 & ResNet-50 v1 \cite{he2016deep} & \cite{resnet2020cifar} & 128 & 200 & 93.9\% & [2K, 25K] & (25k, 50k) & 50K  \\
\hline
\end{tabular}
\label{table:baseline_application}
\end{table}

\subsection{Huge Batch\label{sec:huge_batch}}
%\textcolor{blue}{For ImageNet dataset, we pick a batch size of 819200 and use 512 TPU-v3 chips to train it.}
Researchers are able to scale the batch size of ImageNet/ResNet-50 to 64K and achieve 75.9\% top-1 accuracy in 90 epochs \cite{kumar2019scale}.
As far as we know, no one reported that the batch size of ImageNet/ResNet-50 can be scaled to over 64K without losing accuracy.
%We scale the batch size to 819200, which is a huge batch.
The baseline of our implementation used Momentum SGD optimizer.
The training diverges and only achieves a top-1 accuracy of 0.0977\% for a batch size of 819200.
We train it by 141 steps, which has the same number of epochs as the small-batch baseline. We investigate the effectiveness of existing large batch optimization techniques in the huge batch regime, starting from simple Momentum SGD to more advanced techniques proposed by recent works~\cite{pereyra2017regularizing,kumar2019scale,ying2018image,goyal2017accurate,hoffer2017train}:

\begin{itemize}
    \item {\bf Optimization 0}: this is our baseline; we use Momentum SGD optimizer with linear learning rate scaling (or sqrt scaling if it is better) \cite{krizhevsky2014one}, cosine learning rate decay \cite{loshchilov2016sgdr} (see Appendix), gradient clipping \cite{pascanu2013difficulty} and ghost batch normalization \cite{hoffer2017train}. The top-1 accuracy is 0.0977\%.
    \item {\bf Optimization 1}: we add learning rate warmup \cite{goyal2017accurate}, which allows us to use linear learning rate scaling without diverging. We improved the top-1 accuracy from 0.0977\% to 2.712\%.
    \item {\bf Optimization 2}: we tune learning rate and warmup epochs by an auto-tuner. But we found just using grid-search tuning can roughly get the same accuracy as the auto-tuner in a reasonable computing budget. We improved the top-1 accuracy from 2.712\% to 4.563\%.
    \item {\bf Optimization 3}: we add labeling smoothing \cite{pereyra2017regularizing}, which is a useful technique in large-batch training. The accuracy did not increase and also did not significantly decrease. The accuracy was changed from 4.563\% to 4.226\%.
    \item {\bf Optimization 4}: we switch the optimizer from Momentum SGD to LARS. In this way, we can increase the accuracy from 4.226\% to 18.95\%. 
\end{itemize}
%, which helped researchers achieve state-of-the-art accuracy in large-batch training \cite{kumar2019scale} \cite{ying2018image}

\begin{figure}
\centering
\includegraphics[width=3.9in]{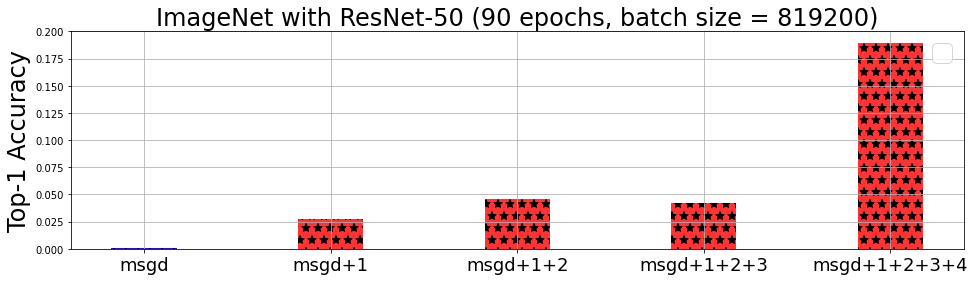}
\caption{msgd: Mementum SGD optimizer with linear learning rate (LR) scaling, cosine LR decay, gradient clipping and ghost batch norm; 1: LR warmup; 2: tuning LR and warmup epochs; 3: label smoothing; 4: LARS. This figure shows the top1 accuracy for huge-batch Imagenet training.}
\label{fig:huge_resnet50_optimization}
\end{figure}

\begin{figure}[htbp]
\centering
\subfigure[Momentum SGD]{
\begin{minipage}[t]{0.5\linewidth}
\centering
    \includegraphics[width=0.87\linewidth]{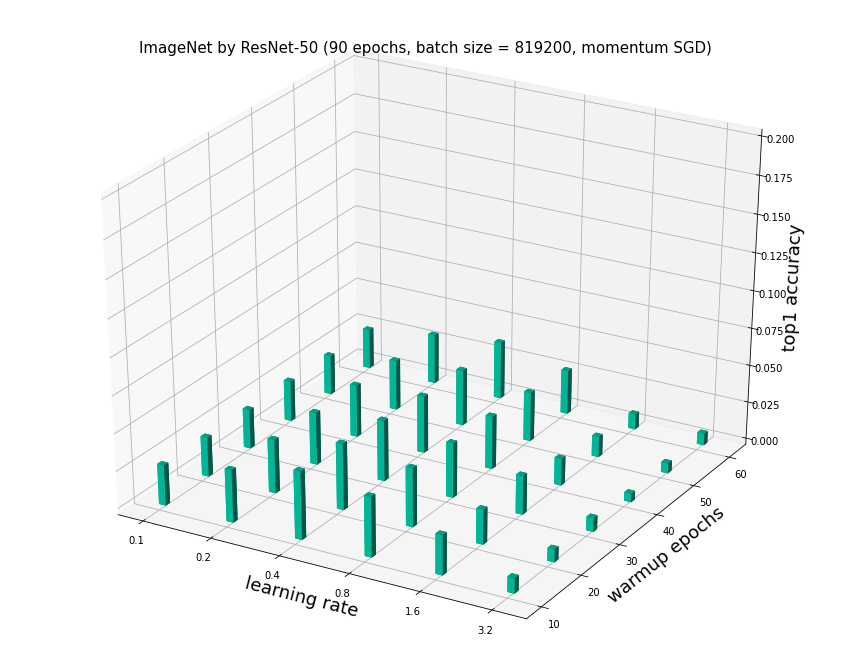}
\end{minipage}%
}%
\subfigure[LARS]{
\begin{minipage}[t]{0.5\linewidth}
\centering
    \includegraphics[width=0.87\linewidth]{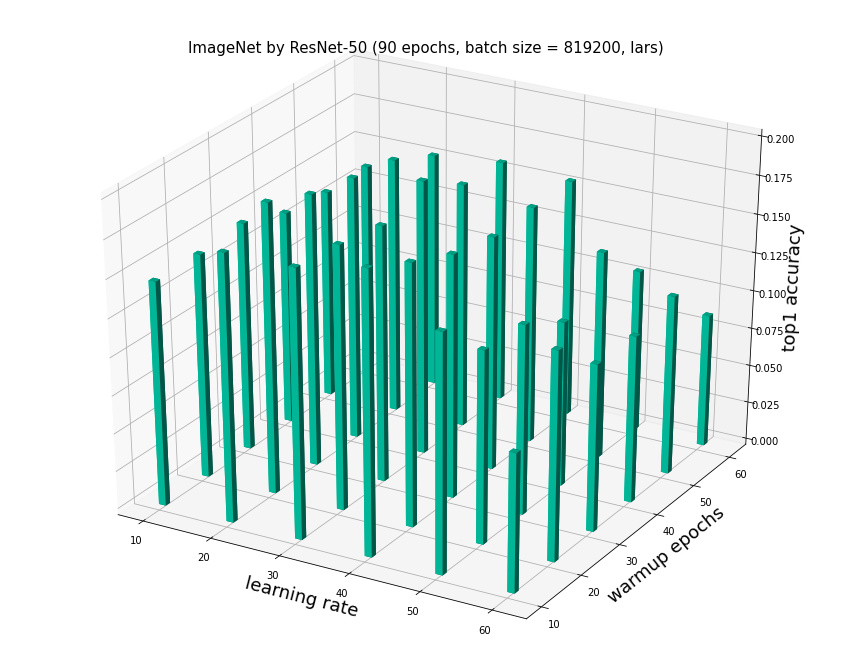}
%\caption{fig2}
\end{minipage}
}%
\centering
\caption{The results of ImageNet training by ResNet-50 (90 epochs or 141 iterations, batch size = 819200). We only show the two most important hyper-parameters and the tuning region with highest accuracy. We can observe that LARS is very necessary in huge-batch learning.}
\label{fig:lars_resnet_huge}
\end{figure}

The results are summarized in Figure \ref{fig:huge_resnet50_optimization}.
Since LARS can give us largest boost in accuracy, we further study its sensitivity to hyperparameters in Figure \ref{fig:lars_resnet_huge}, by varying the two most important hyperparameters, learning rate and number of warmup epochs. From Figure \ref{fig:lars_resnet_huge}, we can observe that the advantage layer-wise adaptive technique provided by LARS optimizer is maintained over a large number of hyperparameters, while the Momentum SGD fails to obtain good accuracy regardless of hyperparameter tuning. We conclude that LARS can make a difference in huge-batch learning.

\begin{figure}
\centering
\includegraphics[width=3.9in]{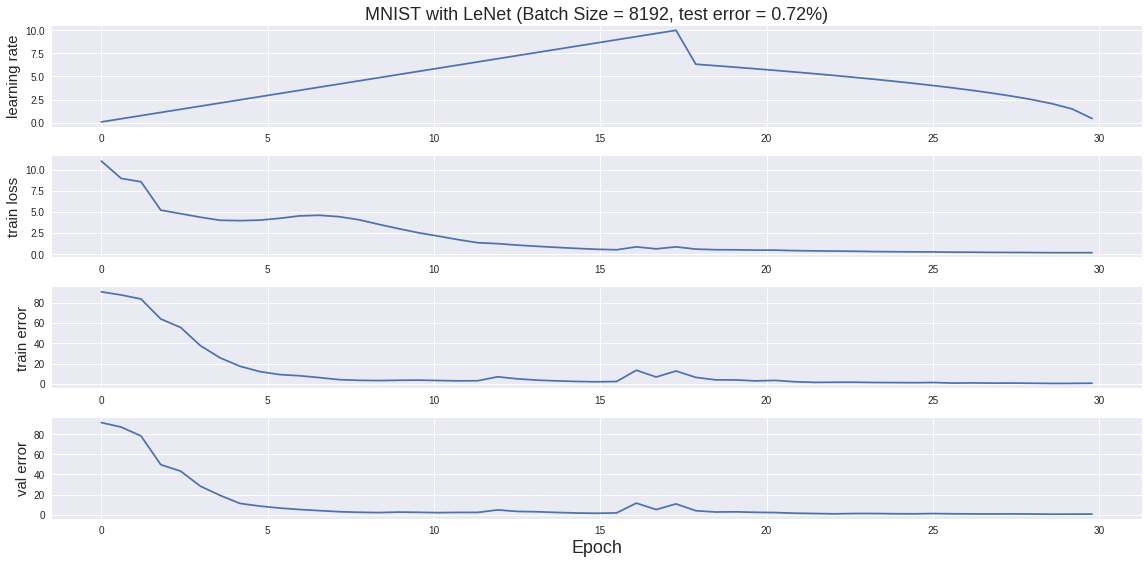}
\caption{LARS optimizer with learning rate warmup and polynomial learning rate decay. The baseline with a batch size of 256 can achieve 99.2\% test accuracy (0.8\% test error rate) in 30 epochs.}
\label{fig:lars_mnist_8k}
\end{figure}

We also study an application with relatively small number of training samples.
We pick MNIST with LeNet.
We partition the dataset as 55K/5K/10K samples for training/validation/testing\footnote{\scriptsize We partition the dataset as 60K/10K samples for training/testing in Table \ref{table:baseline_application}}.
By using Momentum SGD, the baseline can achieve 99.2\% accuracy in 30 epochs with a batch size of 256.
As mentioned in Table \ref{table:baseline_application}, 8K is the boundary between the large batch region and the huge batch region.
When we scale the batch size to 8K, we find the baseline (Momentum SGD) diverges in the training.
So we tried other state-of-the-art optimizers like AdaGrad, RMSprop, Adam and LARS.
AdaGrad and RMSprop can not achieve an accuracy over 99\%.
Adam has a slight accuracy loss.
However, LARS achieved a better accuracy than the baseline.
We want to find the reason behind this.
We find one key feature of LARS is the layer-wise adaptive learning technique.
So we add layer-wise technique to all the other optimizers.
After that, SGD and AdaGrad can get the same accuracy as the baseline.
However, Adam becomes very unstable and RMSprop diverges in the middle of the training.
We find the reason is that the trust ratio\footnote{\scriptsize In the simplest form (for SGD), trust ratio is the ratio between the L2 norm of the weights and the L2 norm of the gradients.} of layer-wise learning is either too large or too small for Adam and RMSprop.
So we add a bound to the trust ratio for Adam optimizer, which is a similar idea as AdaBound \cite{luo2019adaptive}.
By doing so, it can also achieve 99.4\% accuracy in 30 epochs with a batch size of 8K, which is the same as LARS.
It is worth noting that Adam + layer-wise + bound-ratio is a special case of LAMB optimizer \cite{you2019large}.
The results are shown in Table \ref{table:mnist_8k_accuracy}.
Figure \ref{fig:lars_mnist_8k} shows all the training details of LARS optimizer.
%We use learning rate warmup and polynomial decay.

\begin{table}[ht]
\tiny
\renewcommand{\arraystretch}{1.3}
\caption{Scaling the batch size to 8K}
\centering
\begin{tabular}{|c|c|c|c|c|}
\hline
Solver & batch=256 & Original & + layer-wise & + layer-wise and bound-ratio\\
\hline
\hline
SGD & 99.2\% & diverge & 99.2\% & 99.2\%\\
\hline
AdaGrad & 99.2\% & 98.3\% & 99.3\% & 99.3\%\\
\hline
RMSprop & 99.2\% & 98.6\% & diverge & 99.0\%\\
\hline
Adam & 99.3\% & 99.1\% & unstable & 99.4\%\\
\hline
LARS & 99.3\% & 99.4\% & N/A & N/A\\
\hline
\end{tabular}
\label{table:mnist_8k_accuracy}
\end{table}

\begin{figure}
\centering
\includegraphics[width=3.9in]{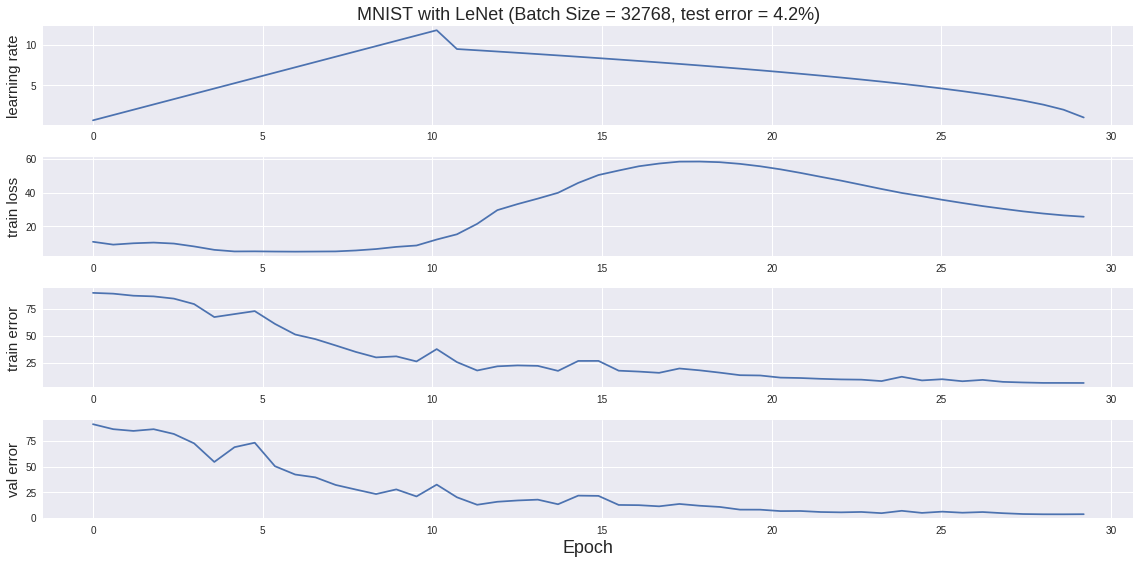}
\caption{LARS optimizer with learning rate (LR) warmup and polynomial LR decay.}
\label{fig:lars_mnist_32k_1}
\end{figure}

\begin{figure}
\centering
\includegraphics[width=3.9in]{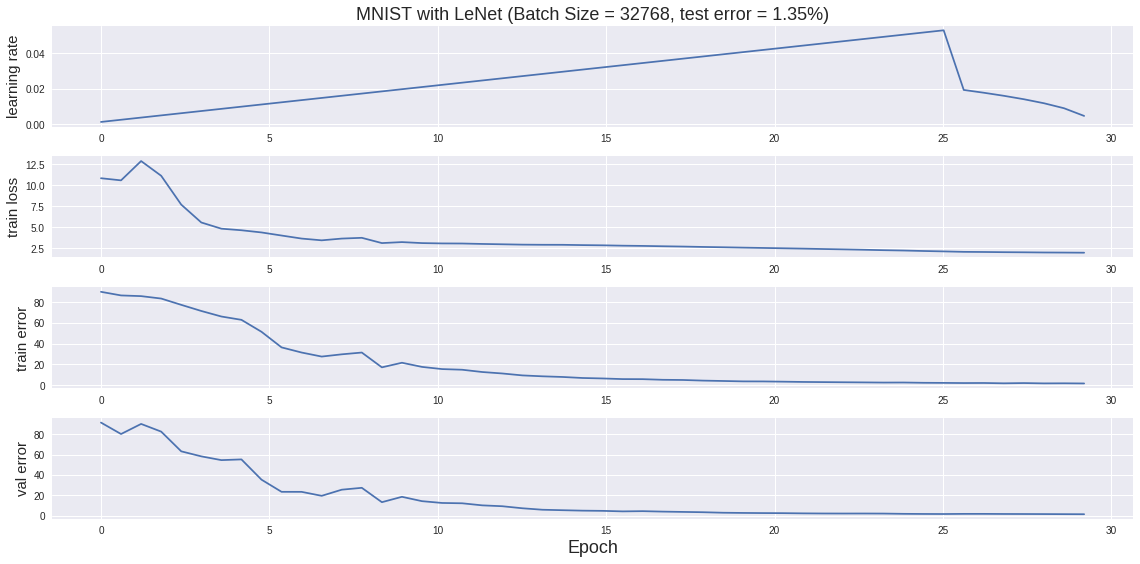}
\caption{LAMB optimizer with extremely long LR warmup epochs and polynomial LR decay. }
\label{fig:lamb_mnist_32768}
\end{figure}

Since LARS with learning rate warmup and polynomial decay gave us best performance for large-batch MNIST training, we use this scheme for huge-batch MNIST training.
We increase the batch size from 8K to 32K.
However, the testing error rate increased from 0.8\% to 4.2\% (Figure \ref{fig:lars_mnist_32k_1}).
If we use other optimizers, the error rate is higher (8\%).
%If we use other optimizers, the model usually loses 7\% of testing accuracy (99.2\% -$>$ 92\%) or diverges.
In Figure \ref{fig:lars_mnist_32k_1}, we can observe huge batch training may lead the optimizer to a wrong path and the model is not optimized well:
in the middle of the training, even though the training error rate and validation error rate are becoming lower, the training loss is increasing. 
% We think the reason is because the optimizer falls in a wrong path in the training and the model actually is not optimized well.
Thus, in this situation, the huge batch training is suffering not only a generalization problem but also an optimization problem.
After trying various different optimization techniques, we find only LAMB optimizer with extremely long learning rate warmup epochs and polynomial learning rate decay can stabilize the training and generalize well (Figure. \ref{fig:lamb_mnist_32768}).
This scheme gets 98.7\% testing accuracy.
However, even using this scheme, we can not reach the target testing accuracy from the baseline -- huge batch-training optimization may have an inherent generalization problem that cannot be resolved by any existing techniques.
Huge-batch ImageNet/ResNet-50 training has a similar problem but is much worse than MNIST/LeNet.
The target top-1 accuracy is 75.9\% in 90 epochs. 
A huge-batch of 819200 only achieves an accuracy lower than 20\%.
We conclude that huge-batch training suffers generalization problems and sometimes also optimization problems with whatever current first-order optimization techniques.
%There are some techniques (e.g. cyclical learning rate) that did not work well.
%We present them in the Appendix.

%The baseline with a batch size of 256 can achieve 99.2\% test accuracy (0.8\% test error rate) in 30 epochs.

\subsection{Full Batch}
\begin{figure}
\centering
\includegraphics[width=3.6in]{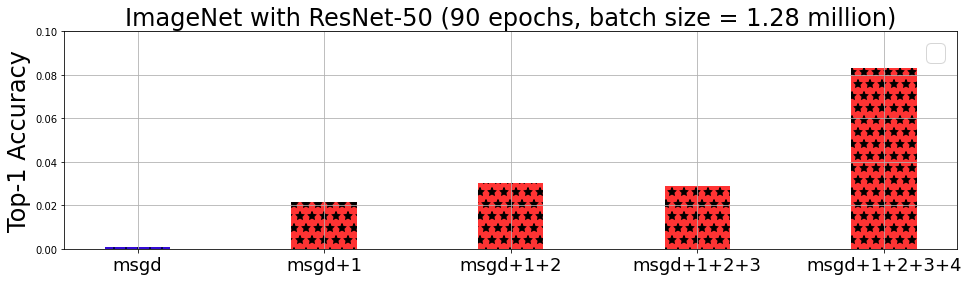}
\caption{msgd: Mementum SGD optimizer with linear learning rate (LR) scaling, cosine LR decay, gradient clipping and ghost batch norm; 1: LR warmup; 2: tuning LR and warmup epochs; 3: label smoothing; 4: LARS. This figure shows the top1 accuracy for full-batch ImageNet training.}
\label{fig:full_batch_resnet50_change}
\end{figure}

\begin{figure}[htbp]
\centering
\subfigure[Momentum SGD]{
\begin{minipage}[t]{0.5\linewidth}
\centering
    \includegraphics[width=0.87\linewidth]{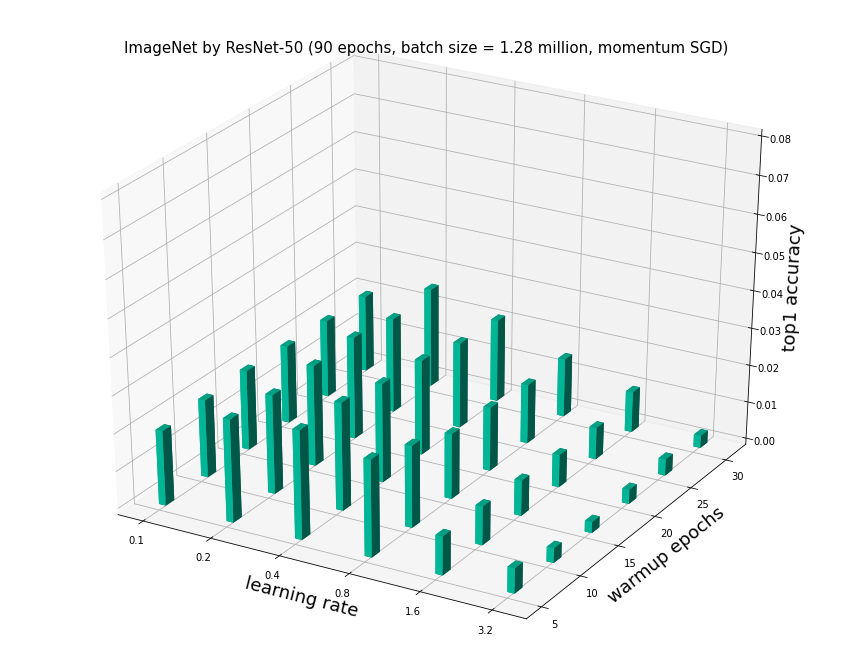}
\end{minipage}%
}%
\subfigure[LARS]{
\begin{minipage}[t]{0.5\linewidth}
\centering
    \includegraphics[width=0.87\linewidth]{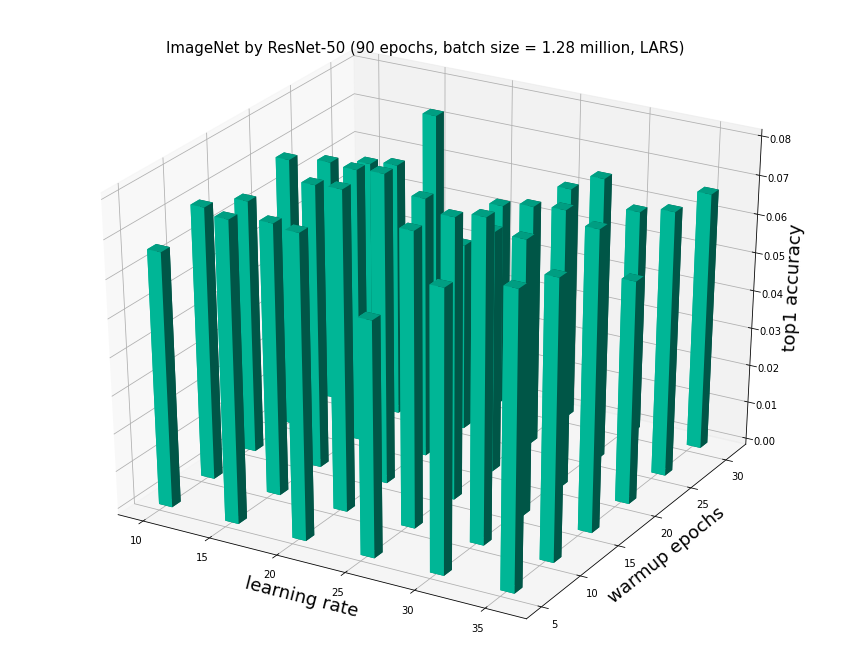}
%\caption{fig2}
\end{minipage}
}%
\centering
\caption{Impact of hyperparameters on ImageNet/ResNet-50 in the full batch regime (90 epochs or 90 iterations, batch size = 1.28 million). We only show the two most important hyper-parameters and the tuning region with highest accuracy. We can see that LARS is very useful in full-batch learning.}
\label{fig:full_batch_resnet_lars_momentum}
\end{figure}

In this section, we increase the batch size from the huge-batch region to the full-batch case.
For ImageNet/ResNet-50, we increase the batch size from 819200 to 1.28 million, which is the first time demonstrated in literature.
For MNIST/LeNet, we partition the dataset as 60K/10K for training/testing. We increase the batch size from 32768 to 60K.
We used the optimization techniques in Section \ref{sec:huge_batch}.
Even though we only increase the batch size by less than two times, we find the accuracy becomes much lower.
The results are shown in Figures \ref{fig:full_batch_resnet50_change} and \ref{fig:full_batch_resnet_lars_momentum}.
We can observe that LARS can also make a significant difference in full-batch training.
However, we can not reach the target testing accuracy in 90 epochs.
%For MNIST dataset, we partition the dataset as 60K/10K samples for training/testing.
%We the batch size to 60K. 
Momentum SGD is our baseline optimizer. LARS is the optimizer achieving the best accuracy for us. 
For MNIST, huge-batch can not reach the 99.2\% target accuracy (results in Appendix). 
LARS can achieve a higher accuracy than Momentum SGD. 
We can also observe that LARS is more stable than Momentum SGD for huge-batch training.

%\begin{figure}
%\includegraphics[width=4.8in]{pic/momentum_mnist.png}
%\caption{\label{fig:momentum_mnist}xxx xxx xxx}
%\end{figure}

%\begin{figure}
%\includegraphics[width=4.8in]{pic/lars_mnist.png}
%\caption{\label{fig:lars_mnist}xxx xxx xxx}
%\end{figure}

\section{Discussion}

\subsection{Train longer, generalize better?\label{sec:discussion1}}
\begin{figure}
\centering
\includegraphics[width=3.6in]{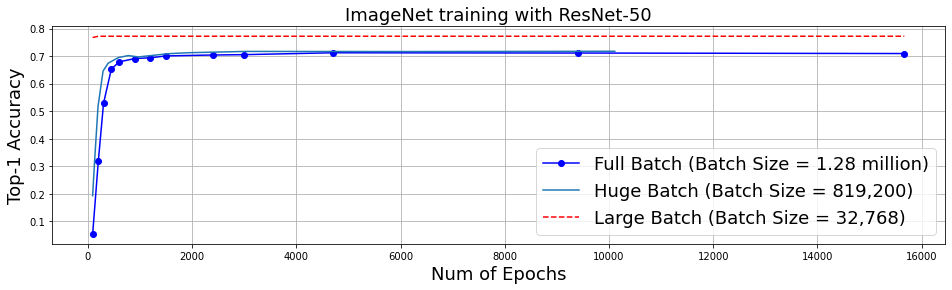}
\caption{For a realistic application like ImageNet/ResNet-50, training longer can not close the generalization gap for huge-batch training and full-batch training. The batch size of 32K is able to reach a testing accuracy of 76.7\% in just 90 epochs. However, even we increase the number of epochs to over 10000, huge-batch and full-batch can not reach the target accuracy.}
\label{fig:epoch_resnet}
\end{figure}

We focus on ImageNet/ResNet-50 training in Sections \ref{sec:discussion1} and \ref{sec:discussion2}.
Our experiments indicate that huge/full-batch training suffers a serious generalization problem.
Hoffer et al. \cite{hoffer2017train} suggest training longer will lead to a better generalization for large-batch training.
We want to study this validity of this suggestion for huge/full-batch training.
The large-batch size of 32K is able to reach a testing accuracy of 76.7\% in just 90 epochs.
The huge-batch (batch size = 819200) training only achieves 19\% accuracy in 90 epochs.
By increasing the number of epochs to 1200, we can reach around 70\% accuracy.
However, even with an extravagant computing budget, we can not reach the target accuracy by just training longer.
Figure \ref{fig:epoch_resnet} shows that there is almost no accuracy improvement as we increase the number of epochs from 3000 to 10000.
For a huge batch size of 819200, the best accuracy we observed is 71.8\%.
For full-batch training, the best accuracy we can get within \emph{15000 epochs} is 71.1\%.
We hypothesize there is an inherent generalization gap in huge/full-batch optimization and simply training longer can not close this gap.

\begin{figure}
\centering
\includegraphics[width=3.6in]{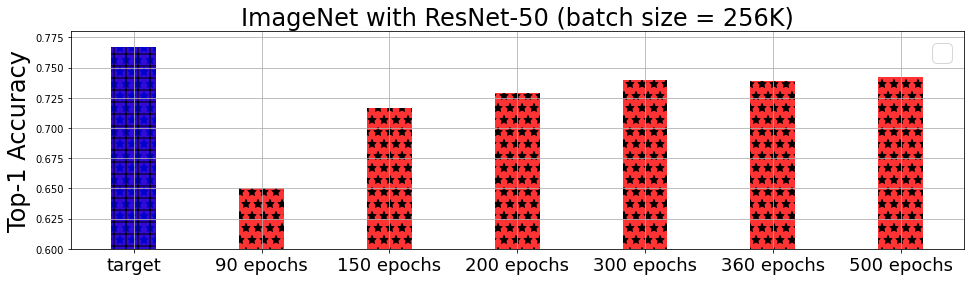}
\caption{Because of distributed batch normalization, the largest useful batch size for the current TPU Pod is 256K. However, we can not reach the target accuracy for ImageNet/ResNet-50.}
\label{fig:lars_resnet50_256k}
\end{figure}

A typical approach on distributed systems is to perform batch normalization (BN) per replica, which reduces the cross-device communication cost.
However, distributed BN may also have an impact on the testing accuracy.
%Let us use ImageNet training with ResNet-50 to illustrate the idea here.
For example, Ying et al. observed using an effective batch size of 64 for batch normalization can lead to the best accuracy (Figure 6 of \cite{ying2018image}).
We use the same approach as Ying et al. \cite{ying2018image}.
Ghost Batch Normalization by Hoffer et al. \cite{hoffer2017train} is a similar idea.
We only conduct reduction across a few peers to compute the mean and variance over a subset of all the replicas.
In our experiments, we observed that we achieve best accuracy on 256 v3 TPU cores for the batch size of 32K.
So the ratio between batch size and the number of cores is 128.
We keep this ratio as we increase the number of cores.
The current largest TPU-based supercomputer has 2048 cores, which corresponds to a batch size of 256K.
That means the largest useful batch size for Ghost Batch Norm on current TPU supercomputer is 256K.
However, we believe 256K is in the huge batch region.
We only achieve 65\% accuracy in 90 epochs.
Even if we increase the number of epochs from 90 to 500, we can not reach the target accuracy (Figure \ref{fig:lars_resnet50_256k}).
This means the current optimization algorithm is not scalable enough to make full use of the current hardware.
To make better use of future hardware, we offer advice for system/hardware designers in the Appendix.

\subsection{Optimization and Generalization\label{sec:discussion2}}

\begin{figure}
\centering
\includegraphics[width=3.9in]{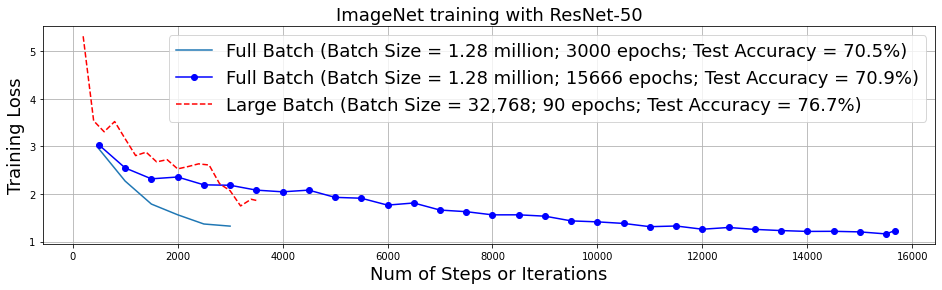}
\caption{Training longer does not generalize better.}
\label{fig:resnet50_loss_accuracy}
\end{figure}

Hoffer et al. \cite{hoffer2017train} introduced ``ultra-slow diffusion'' theory to explain the generalization of large-batch training.
According to some statistical physics models, even though the shape of the
loss function can not be visualized, we can describe the complicated DNN learning process as a random process with some potential.
Specially, they use ``Random Walk on a Random Potential'' to model this complicated optimization and generalization problem. Based on a series of theoretical works \cite{bouchaud1990anomalous, bray2007statistics, dauphin2014identifying, soudry2017exponentially}, they built a relationship between the number of iterations $t$ and the weight at $t$-th iteration $w_t$: $E[||w_t - w_0||^2] \sim (\log{t})^{\frac{4}{\alpha}}$.
It is worth noting that the typical relationship in standard diffusion (on a flat potential) is $||w_t - w_0|| \sim \sqrt{t}$.
They speculate $\alpha$ to be 2, which means the distance between weight $w_t$ with the initial weight $w_0$ increases logarithmically with the number of iterations.
They reached an argument that
%let us assume flat minimum is valuable for improving the generalization performance \cite{keskar2016large}.
the optimizer needs to travel at least a distance of $d$ to find a minimum with the width of $d$, which takes $e^d$ iterations.
%\textcolor{blue}{This leads to the following informal argument (which assumes flat minima are indeed important for generalization). During the initial training phase, to reach a minima of "width" d the weight vector $w_t$ has to travel at least a distance d, and this takes a long time – about $e^d$ iterations. Thus, to reach wide ("flat") minima we need to have the highest possible diffusion rates (which do not result in numerical instability) and a large number of training iterations.}
Essentially, their conclusion implies that given the best optimization technique, the quality of a minimizer is dependent on the number of iterations regardless of the batch size. Thus, the authors proposed an approach, ``regime adaptation'', which suggests using the same number of iterations as the baseline (e.g. batch size = 256) regardless of the batch size.
%\textcolor{blue}{Regime adaptation: Using the tuned learning rate as well as ghost batch-norm, but with an adapted training regime. The training regime is modified to have the same number of iterations for each batch size used - effectively multiplying the number of epochs by the relative size of the large batch.}
%On the other hand, several followup works \cite{goyal2017accurate, ying2018image, kumar2019scale, you2019large} already shown that we can reach the target accuracy in fixed number of epochs for batch size below 128K.
%That means we can linearly reduce the number of iterations as we increase the batch size in a certain range.
However, our results show that huge/full-batch training can not reach the target accuracy even with a large number of epochs.
A natural question to ask is ``why can longer training not lead to a better generalization for huge/full-batch regime?''
%Thus, we try increasing the number of epochs to close this generalization gap.
%The results of ImageNet/ResNet-50 is shown in 
Unfortunately, the relationship between optimization and generalization is missing in their theory. Our experiments (Figure \ref{fig:resnet50_loss_accuracy}) show that training longer leads to a better training loss; however, it can not reach the target testing accuracy from the baseline.
By running the same number of iterations, full-batch can achieve an even lower training loss than the 32K batch size.
%Our experiments have a different conclusion with \cite{hoffer2017train} after we scale the batch size to 819K.
%\textcolor{blue}{However, simple applications like MNIST with LeNet can get the same accuracy by training longer.}
%The tasks of CIFAR-10 and MNIST are easier, but experiments show similar results (Figures \ref{fig:full_batch_cifar_lars_50k_epochs}, \ref{fig:huge_batch_mnist_acc}, and \ref{fig:full_batch_mnist_acc}.).
It indicates there is an inherent generalization gap in huge/full-batch optimization that cannot be closed by any of existing techniques, which should be kept in mind by hardware designers when designing systems for machine learning.
%We propose an informal argument: in the huge-batch regime, the optimizer will need travel at least a distance of $\Theta(Bd)$ to find a minimum with the width of $d$, which takes $e^{Bd}$ iterations.
\section{Conclusion}

%We study the huge/full-batch optimization for deep learning.
%We achieve state-of-the-art results for huge/full-batch optimization.
%We find there is an inherent generalization gap in this problem.
%The ultra-slow diffusion theory can not close this gap.
%We propose three conjectures: (1) for a realistic optimization task (e.g. ImageNet/ResNet-50), there exists a huge-batch range. (2) there is boundary between large-batch and huge-batch. Above this boundary, the learning process becomes totally different. (3) For a realistic dataset/model with full-batch size, we can not reach the target accuracy in polynomial time by current optimization techniques.
%We leave the proof of these three conjectures as our future work.

We study the limit of the batch size for deep neural network training.
For the first time we scale the batch size on ImageNet to at least a magnitude larger that all previous works (819200 versus 64K), and provide detailed studies on the performance of many state-of-the-art optimization schemes under this setting. We propose an optimization recipe that is able to improve the top-1 test accuracy by 18\% compared to the baseline.
We identify a \emph{``huge-batch'' regime} for realistic optimization tasks (e.g. ImageNet/ResNet-50) , where the optimization process becomes intrinsically harder and we cannot reach the target test accuracy in polynomial time by applying any of current optimization techniques (empirical observation). The ``ultra-slow diffusion'' theory \cite{hoffer2017train} does not hold any more in this regime.
Our results help system and algorithm designers to understand the limit of parallelism in realistic machine learning tasks and design future machine learning hardware/algorithms accordingly.
For more results and explanations, please check out the Appendix.

\clearpage

\section*{Broader Impact}

We study the potential applications of huge-batch training and full-batch training, which may have an influence on future hardware design.

\bibliographystyle{plain}
\bibliography{references}

\clearpage

\section{Appendix}

\subsection{Discussion on Ultra-Slow Diffusion Theory}
As Section 2 in the main text, let us refer to $g_t = \frac{1}{|\mathcal{S}_t|} \sum_{s_t \in \mathcal{S}_t} \nabla \ell(x_t, s_t)$ as the gradient at $t$-th iteration and $B$ as the batch size.
Here, we define signal as the update in the direction of the true gradient ($g_t^{\parallel}$) and noise as the update perpendicular to the true gradient ($g_t^{\perp}$).
We also define signal-to-noise-ratio as the ratio between the expected power of the signal and the expected power of the noise: $\frac{E(\|g_t^{\parallel}\|_2)}{E(\|g_t^{\perp}\|_2)}$.

\begin{figure}
\centering
\includegraphics[width=4.8in]{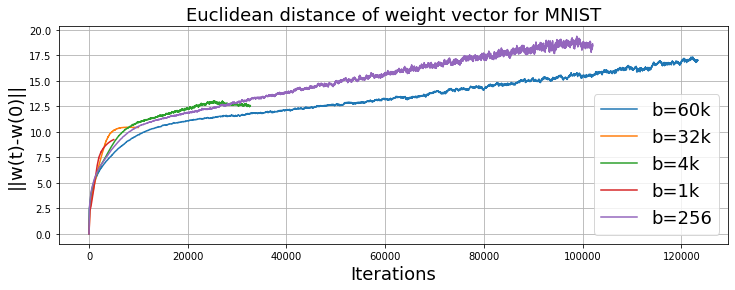}
\caption{Euclidean distance of weight vector from initialization. b denotes the batch size.}
\label{fig:mnist_distance}
\end{figure}

\begin{figure}
\centering
\includegraphics[width=4.8in]{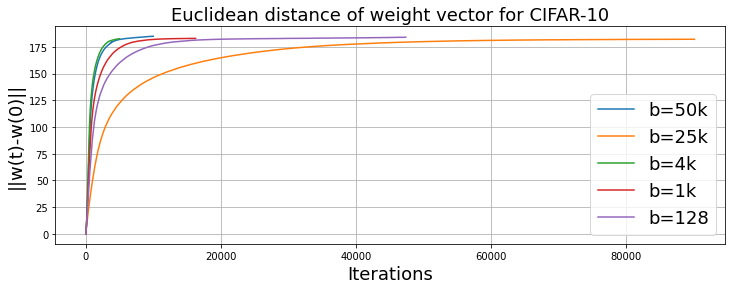}
\caption{Euclidean distance of weight vector from initialization. b denotes the batch size.}
\label{fig:cifar10_distance}
\end{figure}

In this section we want to briefly discuss why training longer can not close the generalization gap in the huge-batch training and full-batch training regimes. In the ``ultra-slow diffusion theory'' \cite{hoffer2017train}, the relationship between the random walk distance and the number of iterations is technically sound and empirically supported by some strong experimental results.
However, we believe the ineffectiveness in huge and full-batch regimes is caused by a few potential mismatches between this theory and the huge/full-batch setting:

\begin{itemize}
    \item Based on the analysis of \cite{keskar2016large}, we assume a minima with a larger "width" will have a better generalization performance. However, it is not guaranteed that we can find a minima of "width" $d$ after traveling a distance of $d$ (although we at least need to travel a distance of $d$ to  find a minima of "width" $d$). Figures \ref{fig:mnist_distance} and \ref{fig:cifar10_distance} support our conjecture. Even all the cases travel the same distance in Figure \ref{fig:mnist_distance} or Figure \ref{fig:cifar10_distance}, each one reaches a very different minima. Figure 2 in \cite{hoffer2017train} only shows that the batch size up to 2K. However, the optimization problem is significantly different if we enter the huge-batch regime.
    \item They did not analyze the effect of signal-to-noise-ratio in stochastic gradients. The signal is increasing linearly with the batch size ($g_t^{\parallel}$ = $\Theta(B)$). The noise is increasing at a sqrt rate with the batch size ($g_t^{\perp}$ = $\Theta(\sqrt{B})$). Thus, increasing the batch size will increase the signal-to-noise-ratio in stochastic gradients, and the full batch setting has no noises. The authors essentially normalized the co-variance in the paper \cite{hoffer2017train}. However, they did not consider the effect of signal-to-noise-ratio in large-batch training. The effect is even bigger in huge-batch training. When the noise is small comparing to the signal, it is unsure if the ``slow diffusion theory'' still holds.
    \item In a real-world training process, the randomness of stochastic potential is very limited. Under the situation of limited training samples and data augmentation, the information is largely redundant and correlated after enough iterations for huge-batch training, violating the assumption of a random potential.
\end{itemize}

\subsection{Additional Results}
\begin{figure}[htbp]
    \subfigure{

    \begin{minipage}[t]{0.4\linewidth}
    
    \includegraphics[width=2.4\linewidth]{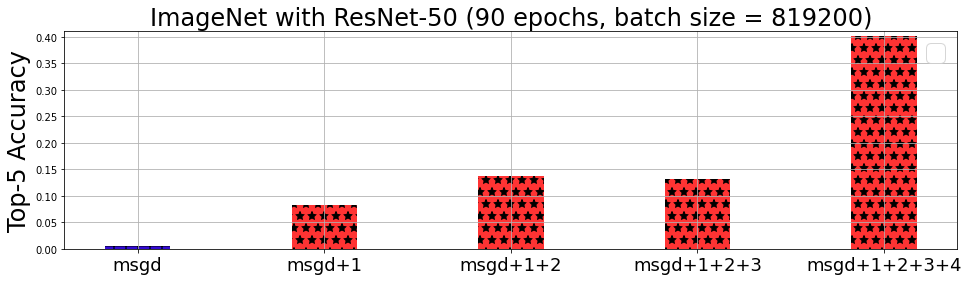}
    %\caption{fig2}
    \end{minipage}%
    }%
    \\
    \subfigure{

    \begin{minipage}[t]{0.4\linewidth}
    
    \includegraphics[width=2.4\linewidth]{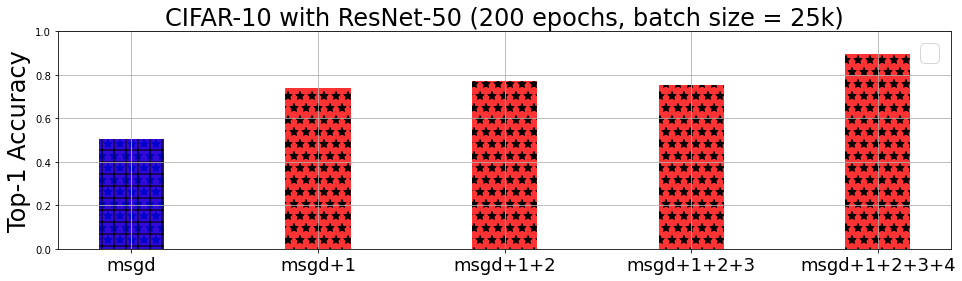}
    \end{minipage}%
    }%
    \caption{msgd: mementum sgd optimizer; 1: learning rate warmup; 2: tuning learning rate and tuning warmup epochs; 3: label smoothing; 4: LARS. This figure shows the top5 accuracy for huge-batch ImageNet training (batch size is 819200, which is larger than 50\% of the total training data size). The batch size of CIFAR-10 is 25K, which is at the boundary between large-batch regime and huge-batch regime.}
    \label{fig:huge_resnet50_optimization_2}
    \end{figure}

We do not have enough space in the main text, so we present additional results here.
Figure \ref{fig:huge_resnet50_optimization_2} shows the top-5 accuracy for huge-batch ImageNet training.
We can observe that LARS is very import for huge-batch training, which in line with the results of previous large-batch training studies \cite{jia2018highly}  \cite{kumar2019scale} \cite{ying2018image}.
The batch size of CIFAR-10 is 25K, which is at the boundary between large-batch regime and huge-batch regime. LARS can also help in this mixed regime.

\begin{figure}[htbp]
    \subfigure{
    \begin{minipage}[t]{0.4\linewidth}
        \includegraphics[width=2.4\linewidth]{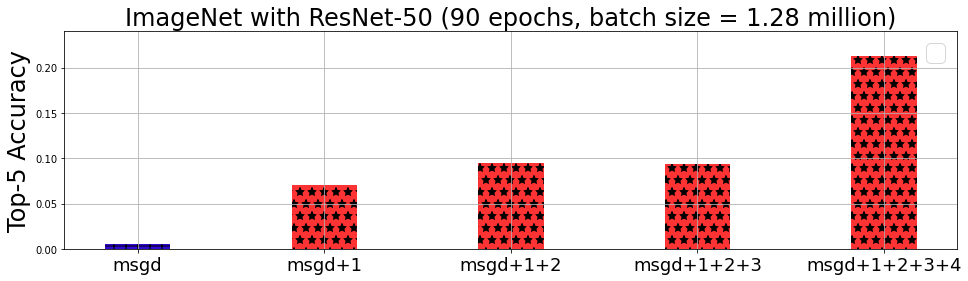}
    %\caption{fig2}
    \end{minipage}%
    }%
    \\
    \subfigure{
    \begin{minipage}[t]{0.4\linewidth}
        \includegraphics[width=2.4\linewidth]{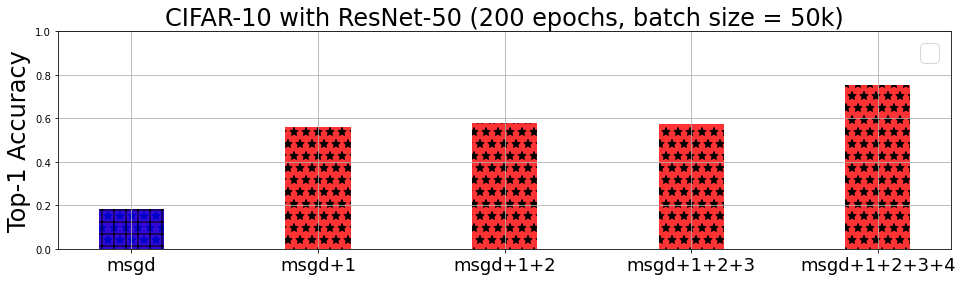}
    %\caption{fig2}
    \end{minipage}%
    }%
    \caption{msgd: mementum sgd optimizer; 1: learning rate warmup; 2: tuning learning rate and tuning warmup epochs; 3: label smoothing; 4: LARS. This figure shows the top5 accuracy for full-batch ImageNet training and the top1 accuracy for full-batch CIFAR-10 training.}
    \label{fig:full_batch_resnet50_change2}
    \end{figure}

\begin{figure}[htbp]
\centering
\subfigure[Momentum SGD on CIFAR-10]{
\begin{minipage}[t]{0.5\linewidth}
\centering
    \includegraphics[width=1.0\linewidth]{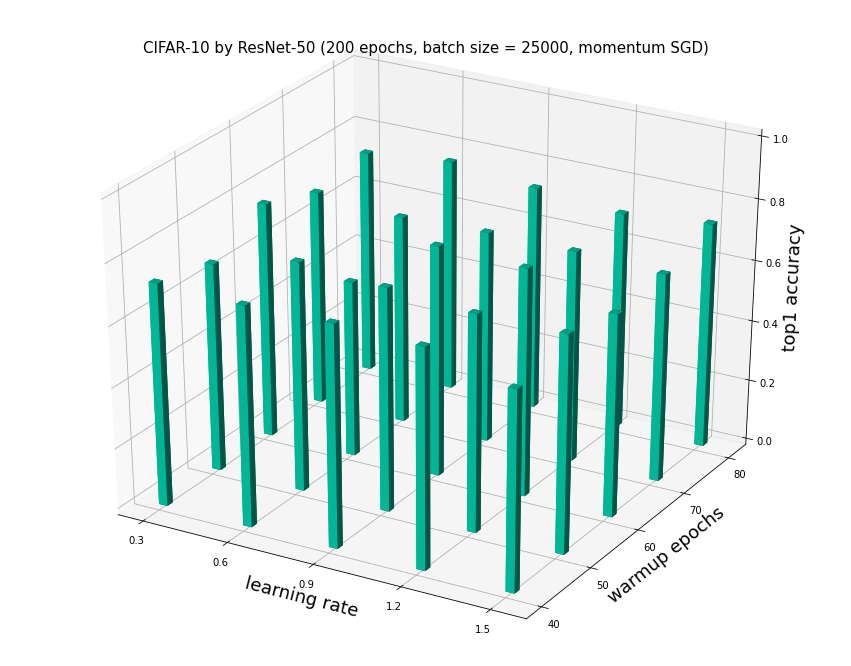}
%\caption{fig2}
\end{minipage}
}%
\subfigure[LARS on CIFAR-10]{
\begin{minipage}[t]{0.5\linewidth}
\centering
    \includegraphics[width=1.0\linewidth]{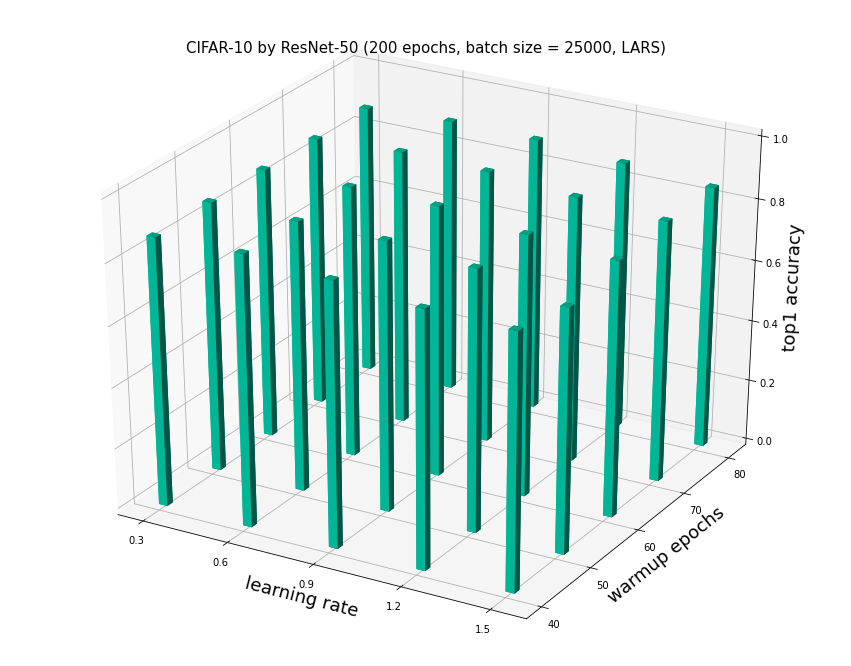}
%\caption{fig2}
\end{minipage}
}%
\centering
\caption{This figure shows the hyper-parameter tuning results of full-batch CIFAR-10 training by ResNet-50 (200 epochs or 200 iterations, batch size = 50K). We only show the tuning region of the best hyper-parameters. We can observe that LARS can increase the accuracy of full-batch training. However, the current first-order optimizers can not achieve the same accuracy as the baseline for full-batch training.}
\label{fig:lars_resnet_huge2}
\end{figure}

\begin{figure}[htbp]
    \subfigure{
    \begin{minipage}[t]{0.4\linewidth}
        \includegraphics[width=2.5\linewidth]{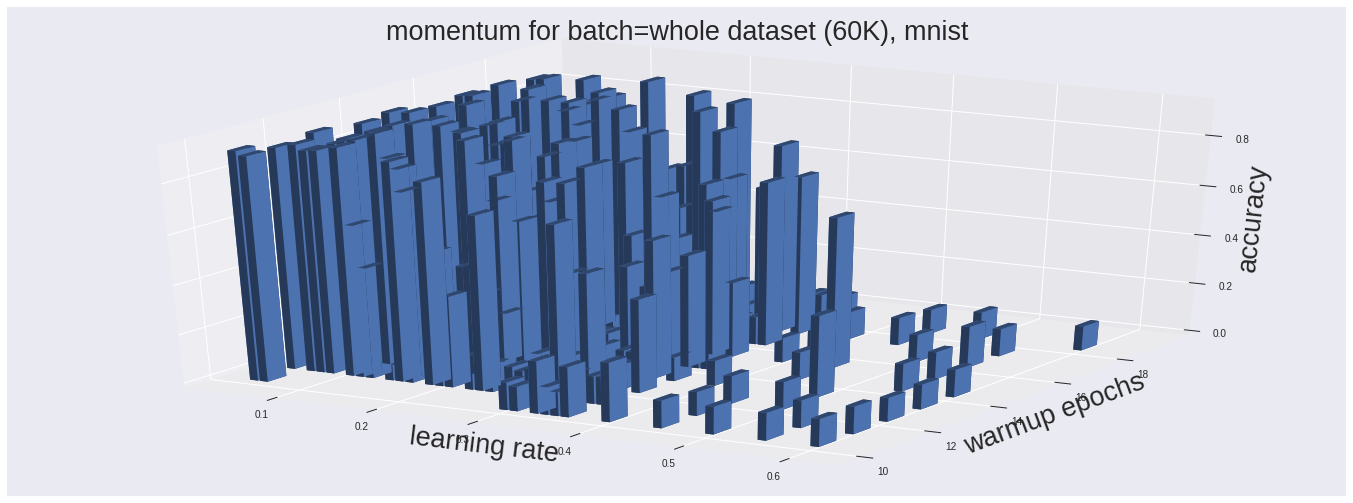}
    \end{minipage}%
    }%
    \\
    \subfigure{
    \begin{minipage}[t]{0.4\linewidth}
        \includegraphics[width=2.5\linewidth]{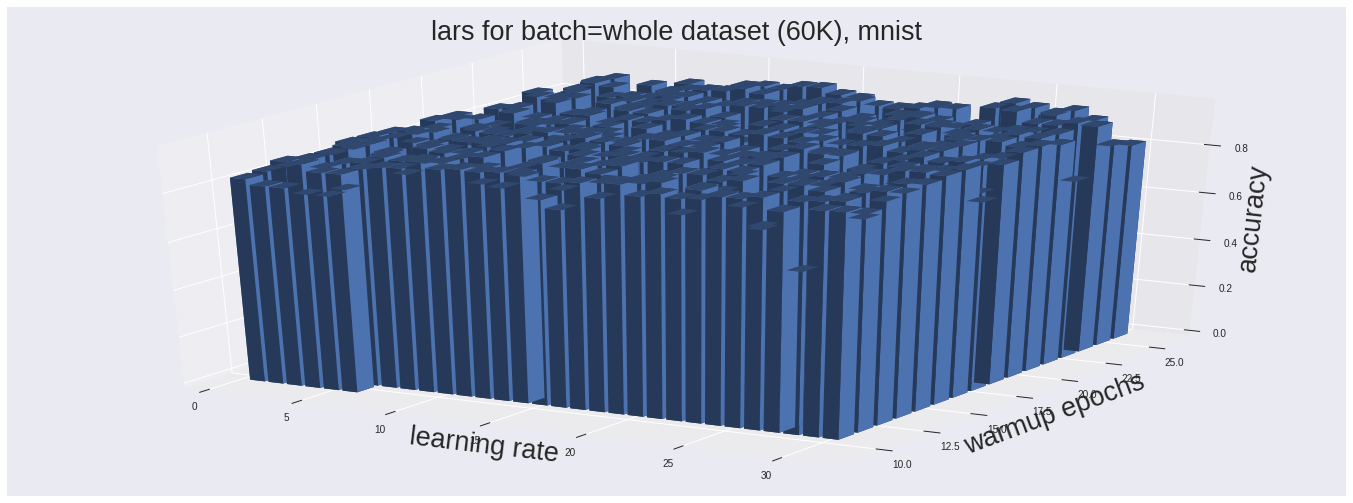}
    %\caption{fig2}
    \end{minipage}%
    }%
    \caption{We partition the dataset as 60K/10K samples for training/testing. We increase the batch size to 60K. Momentum SGD is our baseline optimizer. LARS is the optimizer achieving the best accuracy for us. We can observe that both optimizers can not reach the target accuracy (99.2\%). However, LARS can achieve a higher accuracy than Momentum SGD. This figure only shows the most effective hyper-parameter tuning region. We can also observe that LARS is more stable than Momentum SGD for huge-batch training.}
    \label{fig:full_batch_mnist_lars_mentum}
    \end{figure}

\begin{figure}[htbp]
\centering
\subfigure[Momentum SGD on CIFAR-10]{
\begin{minipage}[t]{0.5\linewidth}
\centering
    \includegraphics[width=1.0\linewidth]{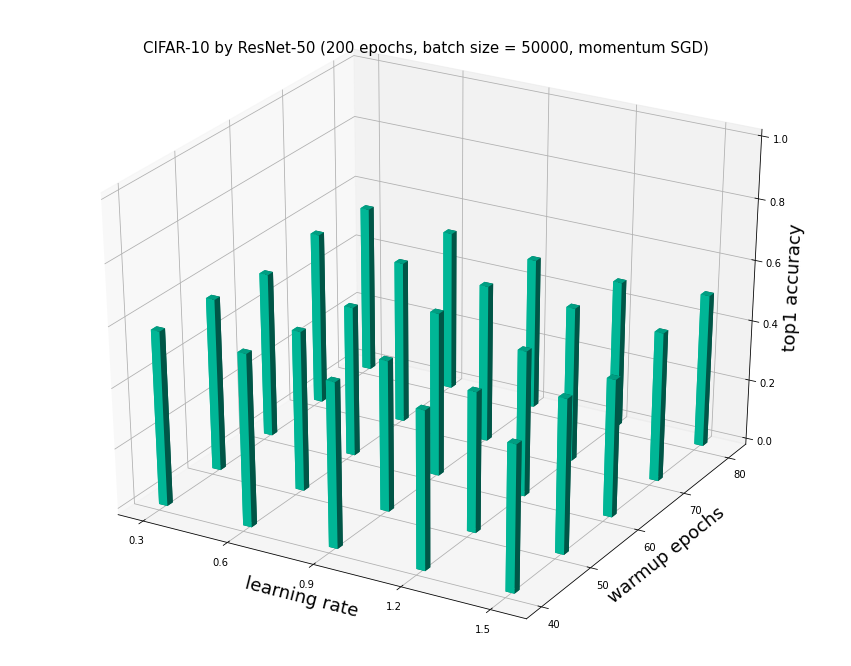}
%\caption{fig2}
\end{minipage}
}%
\subfigure[LARS on CIFAR-10]{
\begin{minipage}[t]{0.5\linewidth}
\centering
    \includegraphics[width=1.0\linewidth]{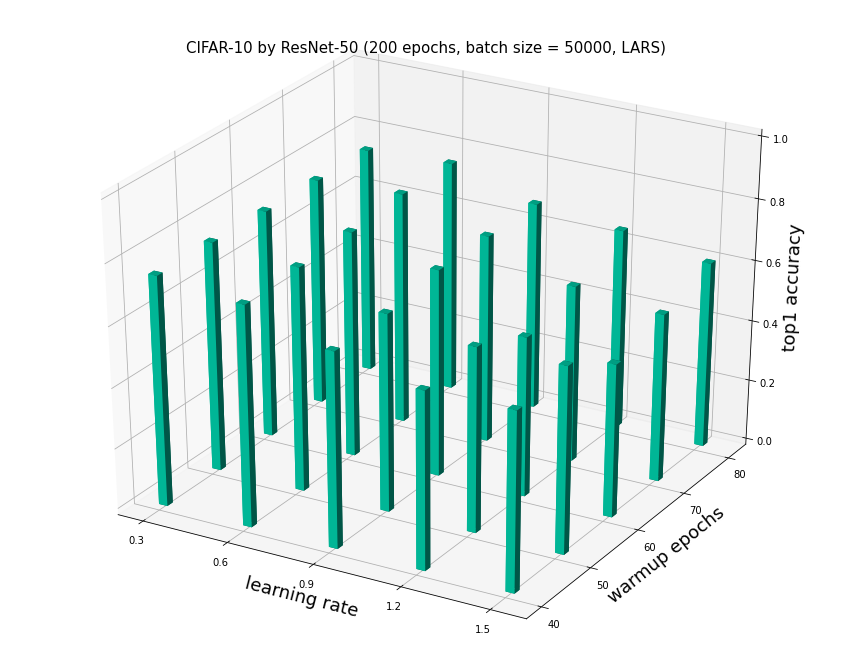}
%\caption{fig2}
\end{minipage}
}%
\centering
\caption{This figure shows the results of ImageNet training by ResNet-50 (90 epochs or 90 iterations, batch size = 1.28 million). We only show the tuning region of the best hyper-parameters. We can observe that the current first-order optimizers can only achieve a very low accuracy for the full-batch training.}
\label{fig:full_batch_resnet_lars_momentum2}
\end{figure}

Figure \ref{fig:full_batch_resnet50_change2} shows the results of full-batch training.
We scale the batch size of ImageNet training to 1.28 million and the batch size of CIFAR-10 to 50K.
It shows the top5 accuracy for full-batch ImageNet training and the top1 accuracy for CIFAR-10 training are increased dramatically after using a series of optimization techniques including LARS.
Figure \ref{fig:lars_resnet_huge2} shows the hyper-parameter tuning results of full-batch CIFAR-10 training by ResNet-50 (200 epochs or 200 iterations, batch size = 50K). We only show the tuning region of the best hyper-parameters. We can observe that LARS can increase the accuracy of full-batch training. However, the current first-order optimizers can not achieve the same accuracy as the baseline for full-batch training.
Figure \ref{fig:full_batch_mnist_lars_mentum} shows the full-batch training results of MNIST.
We partition the dataset as 60K/10K samples for training/testing. 
We increase the batch size to 60K. 
Momentum SGD is our baseline optimizer. LARS is the optimizer achieving the best accuracy for us. We can observe that both optimizers can not reach the target accuracy (99.2\%). However, LARS can achieve a higher accuracy than Momentum SGD. 
Figure \ref{fig:full_batch_mnist_lars_mentum} only shows the most effective hyper-parameter tuning region. We can also observe that LARS is more stable than Momentum SGD for full-batch training.
Figure \ref{fig:full_batch_resnet_lars_momentum2} shows a similar result.

\begin{figure}
\centering
\includegraphics[width=4.8in]{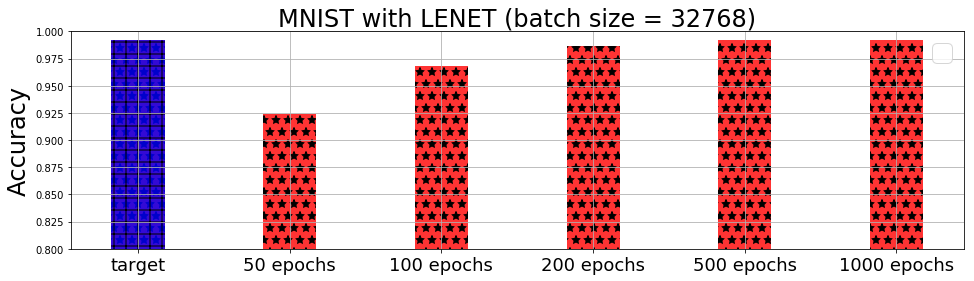}
\caption{MNIST with LeNet (batch size = 32768). The baseline with a batch size of 256 can achieve 99.2\% testing accuracy in 30 epochs. With LARS, the baseline can achieve 99.3\% accuracy in 30 epochs. However, huge-batch can not reach the target accuracy even we train it for a long time. The best accuracy we can get is 99.16\% by 1000 epochs.}
\label{fig:huge_batch_mnist_acc}
\end{figure}

As the results shown in the main text, training longer can not close the generalization gap for huge-batch and full-batch training.
MNIST with LeNet (batch size = 60k). The baseline with a batch size of 256 can achieve 99.2\% testing accuracy in 30 epochs. With LARS, the baseline can achieve 99.3\% accuracy in 30 epochs.
For the huge-batch MNIST training, we set the batch size as 32768.
From Figure \ref{fig:huge_batch_mnist_acc} we can observe that even increase the number of epochs from 30 to 1000, we can only achieve 99.16\% accuracy, which still can not match the target accuracy.
For the full-batch training, the accuracy for 1000-epoch training is even lower, which is only 98.3\%.
It is worth noting that a batch size of 256 can achieve this level of accuracy in just 10 epochs.
That means that there is huge generalization gap between full-batch optimization and regular SGD optimization.
For CIFAR-10 training with ResNet-50 (batch size = 50k), the baseline can achieve 93.9\% accuracy in 200 epochs. However, the full-batch version can not reach the target accuracy even we train it for a long time. The best accuracy we can get is 92.58\% by 5000 epochs (Figure \ref{fig:full_batch_cifar_lars_50k_epochs})).

\begin{figure}
\centering
\includegraphics[width=4.8in]{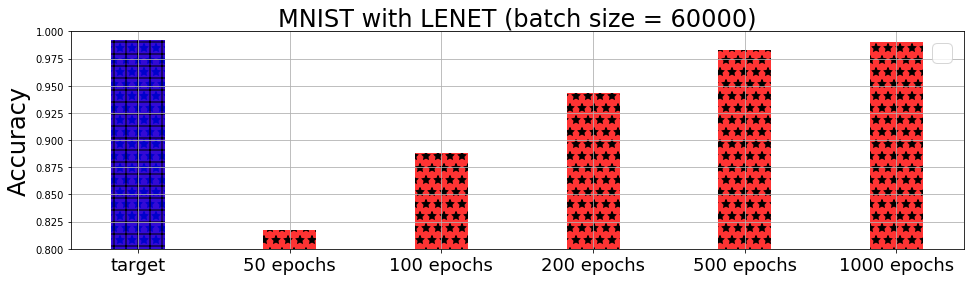}
\caption{MNIST with LeNet (batch size = 60k). The baseline with a batch size of 256 can achieve 99.2\% testing accuracy in 30 epochs. With LARS, the baseline can achieve 99.3\% accuracy in 30 epochs. However, full-batch can not reach the target accuracy even we train it for a long time. The best accuracy we can get is 98.30\% by 1000 epochs.}
\label{fig:full_batch_mnist_acc}
\end{figure}

\begin{figure}
\centering
\includegraphics[width=4.8in]{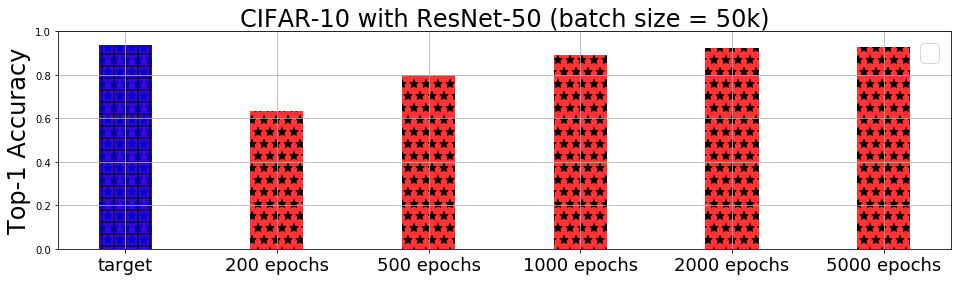}
\caption{CIFAR-10 with ResNet-50 (batch size = 50k). The baseline can achieve 93.9\% accuracy in 200 epochs. However, full-batch can not reach the target accuracy even we train it for a long time. The best accuracy we can get is 92.58\% by 5000 epochs.}
\label{fig:full_batch_cifar_lars_50k_epochs}
\end{figure}

\subsection{Why LARS can help?}
From results in the previous section, we can observe that LARS can make a significant difference in huge-batch training and full-batch training.
We want to briefly explain why LARS can help for huge/full-batch training, which is out of the range of the original LARS paper \cite{you2017scaling}.
As mentioned by Keskar et al. \cite{keskar2016large}, small-batch learner uses noisy gradients in the computation of each step.
The noises actually can push the learner away from the basin of the sharp minimizers.
However, the noises are greatly reduced in large/full batch training, which are not enough to push the learner out of the basin of the sharp minimizers.
Thus, adding noises to huge/full-batch learner may help.
However, how to add proper noises?
We tried adding the Gaussian noises and significantly tuning the hyperparameters, but it did not improve the testing accuracy. 
Specially, we did the following experiments.
\begin{itemize}
    \item Add noise to activations, i.e. the outputs of each layer.
    \item Add noise to weights, i.e. an alternative to the inputs.
    \item Add noise to the gradients, i.e. the direction to update weights.
    \item Add noise to the outputs, i.e. the labels or target variables.
\end{itemize}
Keskar et al. did similar experiments, but it also did not help\footnote{\tiny \url{https://openreview.net/forum?id=H1oyRlYgg&noteId=H1oyRlYgg}}.
On the other hand, LARS computes the trust ratio at each iteration \cite{you2017scaling}:

\begin{equation}
    {r}^l_t = \frac{||w^l_{t-1}||}{||\frac{1}{B} \sum_{i=1}^{B} \nabla f(x_t^i, w^l_{t-1})|| + \lambda ||w^l_{t-1}||} = \frac{||w^l_{t-1}||}{||g^l_t|| + \lambda ||w^l_{t-1}||}
\end{equation}

where $l$ is the layer ID, $B$ is the batch size, $t$ is the iteration ID, and $\lambda$ is the weight decay (e.g. $\lambda$ = 0.01).
Then LARS uses trust ratio to multiply the learning rate.
From Figure 2 of this link\footnote{\tiny https://arxiv.org/pdf/1708.03888v3.pdf} we can observe that: 
\begin{itemize}
    \item For a different iteration, the learning rate is different.
    \item For a different layer, the learning rate is different.
    \item For a different batch size, the learning rate is different.
\end{itemize}
Therefore, we think the trust ratio of LARS can provide dynamics to the learning process.
The dynamics may act as a proper noise to help the learner get out of the sharp minimum.

\subsection{Identify the huge-batch regime}
\begin{figure}
\includegraphics[width=4.8in]{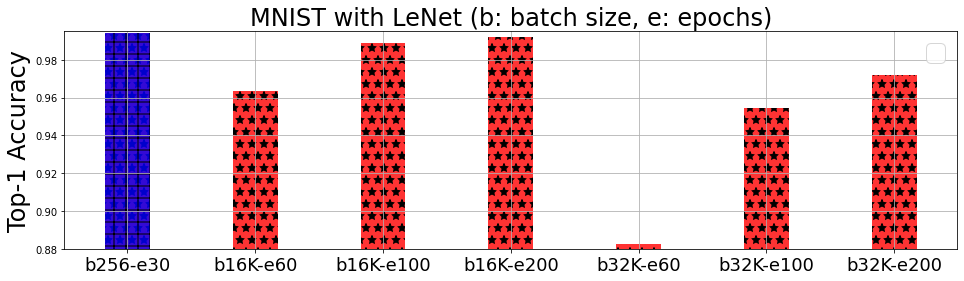}
\caption{Even we increase the number of training epochs from 30 to 200, we can not reach the target accuracy for MNIST/LeNet at a batch size of 16K or larger.}
\label{fig:mnist_lenet_identify}
\end{figure}

\begin{figure}
\includegraphics[width=4.8in]{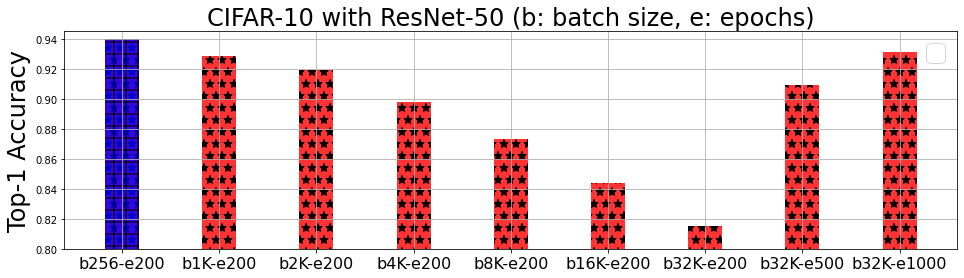}
\caption{Even we increase the number of training epochs from 200 to 1000, we can not reach the target accuracy for CIFAR-10/ResNet-50 at a batch size of 25K or larger.}
\label{fig:cifar10_resnet50_identify}
\end{figure}

As mentioned in the main text, we identified the huge batch regime of ImageNet/ResNet-50 as (64K, 1.28 million).
Kumar et al. \cite{kumar2019scale} reported that they can reach the target accuracy for ImageNet/ResNet-50 at a batch size of 64K by using LARS optimizer \cite{you2017scaling}.
We use the training recipe of Kumar et al. \cite{kumar2019scale} and an auto-tuner to maximize the accuracy of ImageNet/ResNet-50 at a batch size of 128K.
However, we can not reach the target accuracy.
Table \ref{table:resnet_128k_batch} shows the best results that we can achieve.
The best top-1 accuracy we can get is only 73.37\%, which is much lower than the target accuracy (75.9\%).
Thus, we identify a batch size over 64K is huge batch for ImageNet/ResNet-50 training.
Figure \ref{fig:mnist_lenet_identify} shows that we can identify the huge-batch regime of MNIST/LeNet training as (8K, 60K).
Figure \ref{fig:cifar10_resnet50_identify} shows that we can identify the huge-batch regime of CIFAR-10/ResNet-50 training as (25K, 50K).

\begin{table}[ht]
\renewcommand{\arraystretch}{1.3}
\caption{ImageNet training with ResNet-50. The batch size is 128K, weight decay is 0.0001, and label smoothing is 0.1.}
\centering
\begin{tabular}{|c|c|c|c|c|c|}
\hline
LR & warmup epochs & momentum & LR schedule & TPU chips & Top1 accuracy\\
\hline
\hline
26 & 25 & 0.94429 & cosine & 256 & 72.92\%\\
\hline
28 & 25 & 0.94429 & cosine & 256 & 73.37\%\\
\hline
30 & 25 & 0.94429 & cosine & 256 & 73.08\%\\
\hline
32 & 25 & 0.94429 & cosine & 256 & 72.82\%\\
\hline
26 & 25 & 0.94429 & cosine & 512 & 73.14\%\\
\hline
28 & 25 & 0.95 & cosine & 512 & 73.30\%\\
\hline
30 & 25 & 0.95 & cosine & 512 & 73.14\%\\
\hline
32 & 25 & 0.95 & cosine & 512 & 72.82\%\\
\hline
\end{tabular}
\label{table:resnet_128k_batch}
\end{table}

\begin{figure}
\includegraphics[width=4.8in]{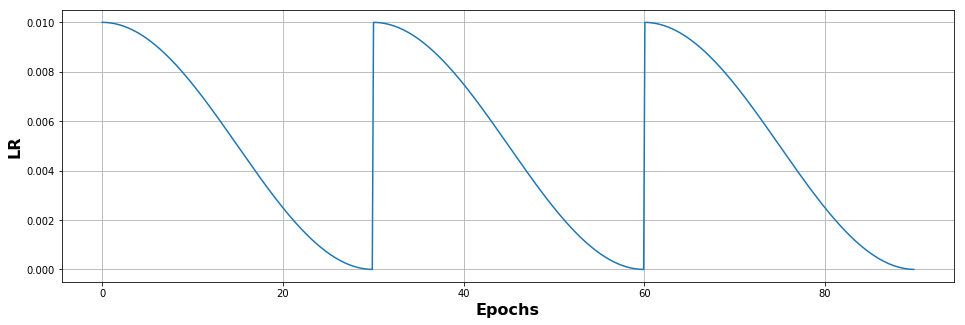}
\caption{Regular cosine learning rate schedule.}
\label{fig:lr_0}
\end{figure}

\begin{figure}
\includegraphics[width=4.8in]{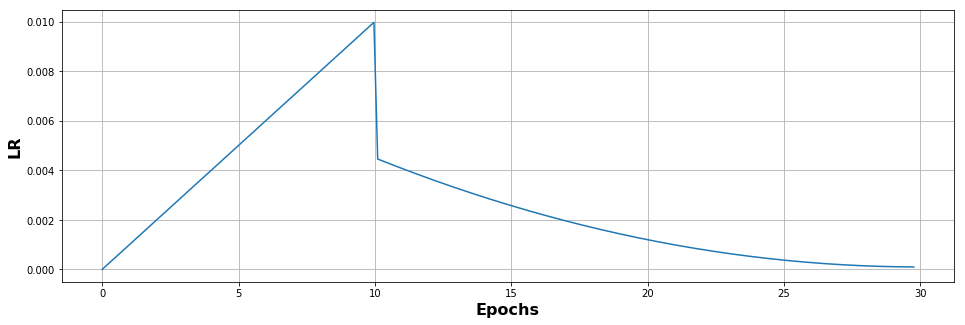}
\caption{Regular learning rate warmup and poly learning rate decay.}
\label{fig:lr_1}
\end{figure}

\begin{figure}
\includegraphics[width=4.8in]{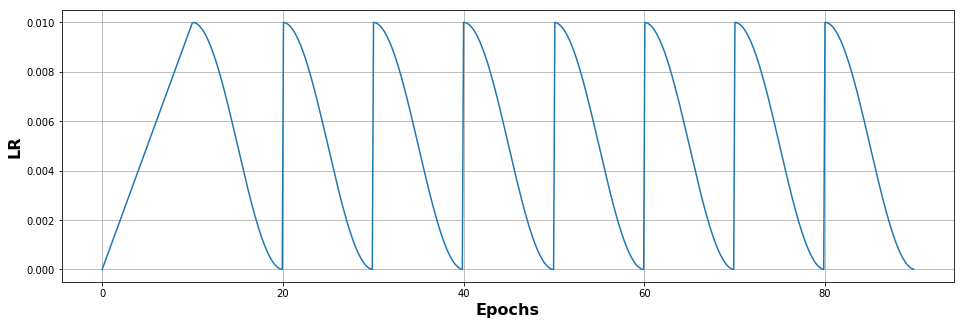}
\caption{Regular learning rate warmup and fine-grained cosine learning rate schedule}
\label{fig:lr_2}
\end{figure}

\begin{figure}
\includegraphics[width=4.8in]{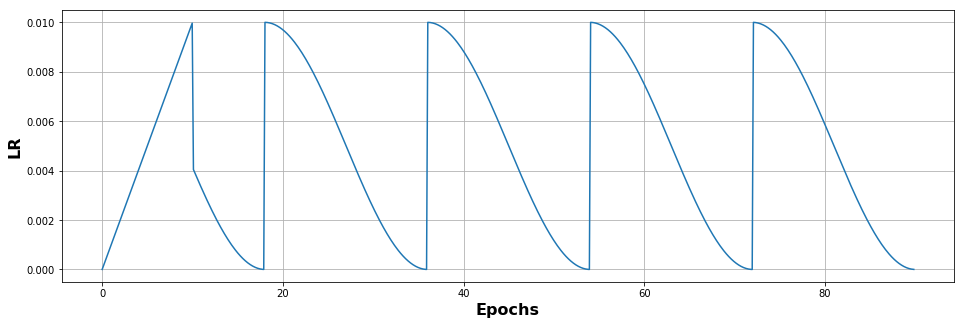}
\caption{Regular learning rate warmup and regular cosine learning rate schedule}
\label{fig:lr_3}
\end{figure}

\begin{figure}
\includegraphics[width=4.8in]{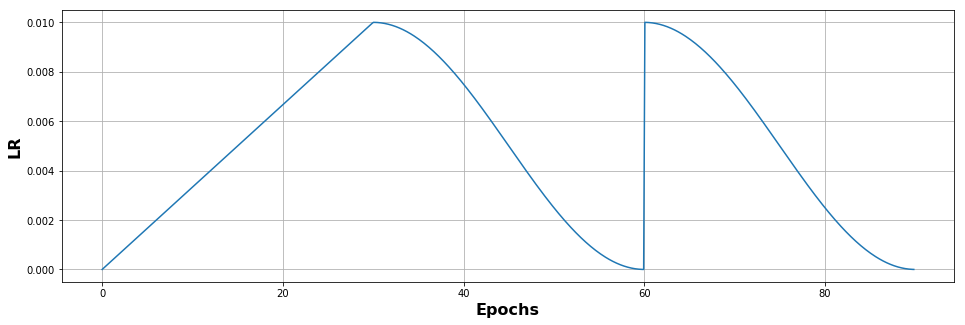}
\caption{Regular learning rate warmup and coarse-grained cosine learning rate schedule}
\label{fig:lr_4}
\end{figure}

\begin{figure}
\includegraphics[width=4.8in]{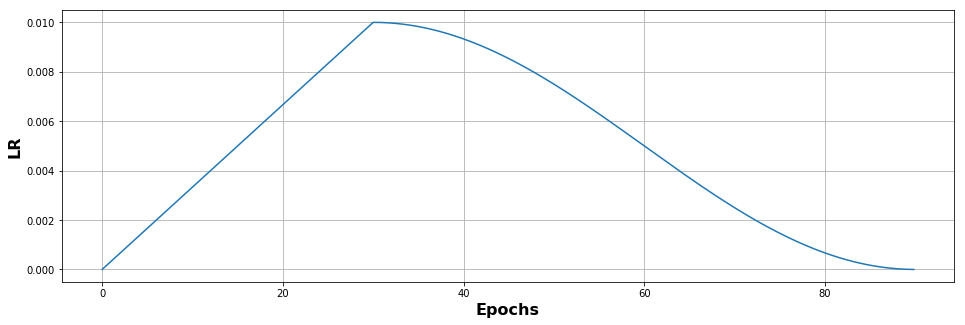}
\caption{Regular learning rate warmup and cosine learning rate decay.}
\label{fig:lr_5}
\end{figure}

\begin{figure}
\includegraphics[width=4.8in]{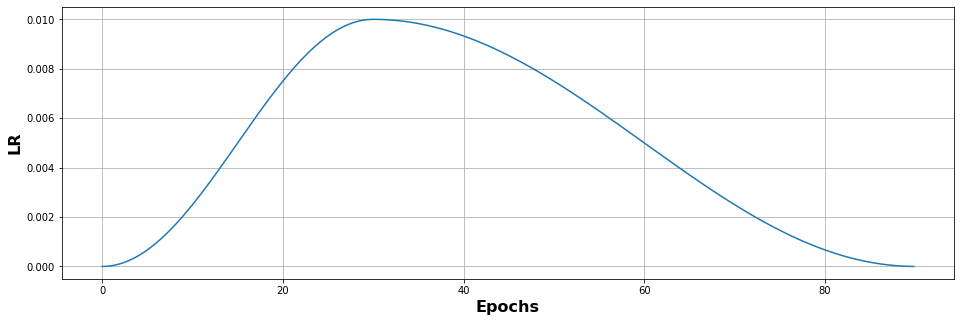}
\caption{Cosine learning rate warmup and cosine learning rate decay.}
\label{fig:lr_6}
\end{figure}

\subsection{Learning Rate Schedule}
Since learning rate is the most important hyper-parameter, we tried several different learning rate schedules.
Specifically, they include regular cosine learning rate schedule (Figure \ref{fig:lr_0}), regular learning rate warmup and poly learning rate decay (Figure \ref{fig:lr_1}), regular learning rate warmup and fine-grained cosine learning rate schedule (Figure \ref{fig:lr_2}), regular learning rate warmup and regular cosine learning rate schedule (Figure \ref{fig:lr_3}), regular learning rate warmup and coarse-grained cosine learning rate schedule (Figure \ref{fig:lr_4}), regular learning rate warmup and cosine learning rate decay (Figure \ref{fig:lr_5}), and cosine learning rate warmup and cosine learning rate decay (Figure \ref{fig:lr_6}).
We pick the best schedule for each application.

For example, we tried cyclical learning rate \cite{smith2017cyclical} scheme in MNIST training.
This brings an additional hyper-parameter: the cyclical length.
If we do not tune it carefully, the testing accuracy can be significantly hurt (error rate increases from 4.2\% to 12.7\% in Figure \ref{fig:lars_mnist_cyc1}).
After tuning it, we can get a slightly better accuracy (error rate decreases from 4.2\% to 3.8\%), which is shown in Figure \ref{fig:lars_mnist_cyc2}.
However, we find in both these two figures, the training loss reached the peaking point in the middle of the training and remain very high in the end of the training.

\begin{figure}
\centering
\includegraphics[width=4.8in]{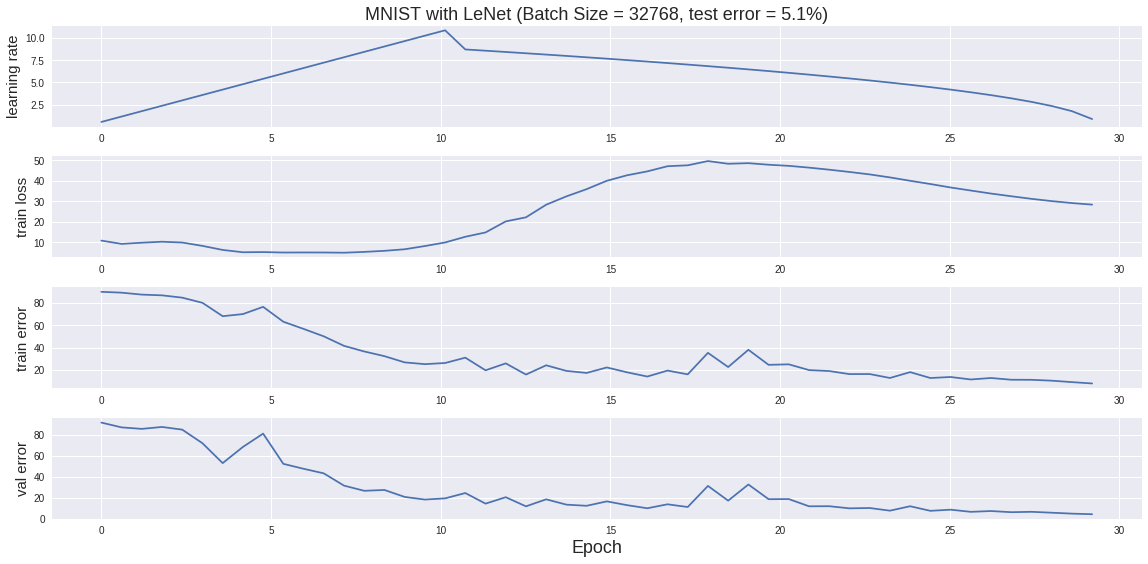}
\caption{\label{fig:lars_mnist_32k_2}LARS optimizer with learning rate warmup and polynomial learning rate decay.}
\end{figure}

\begin{figure}
\centering
\includegraphics[width=4.8in]{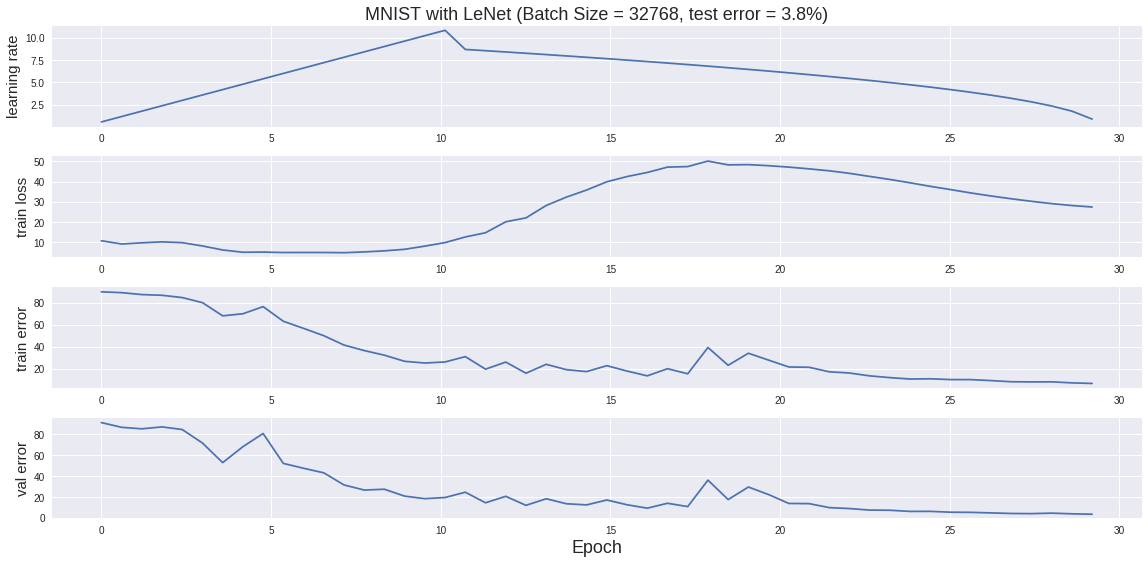}
\caption{LARS optimizer with learning rate warmup and polynomial learning rate decay. After we increase the batch size from 8K to 32K, we observed a significant increase in the test error rate (from 0.8\% to 4.2\%).}
\label{fig:lars_mnist_32k_3}
\end{figure}

\begin{figure}
\centering
\includegraphics[width=4.8in]{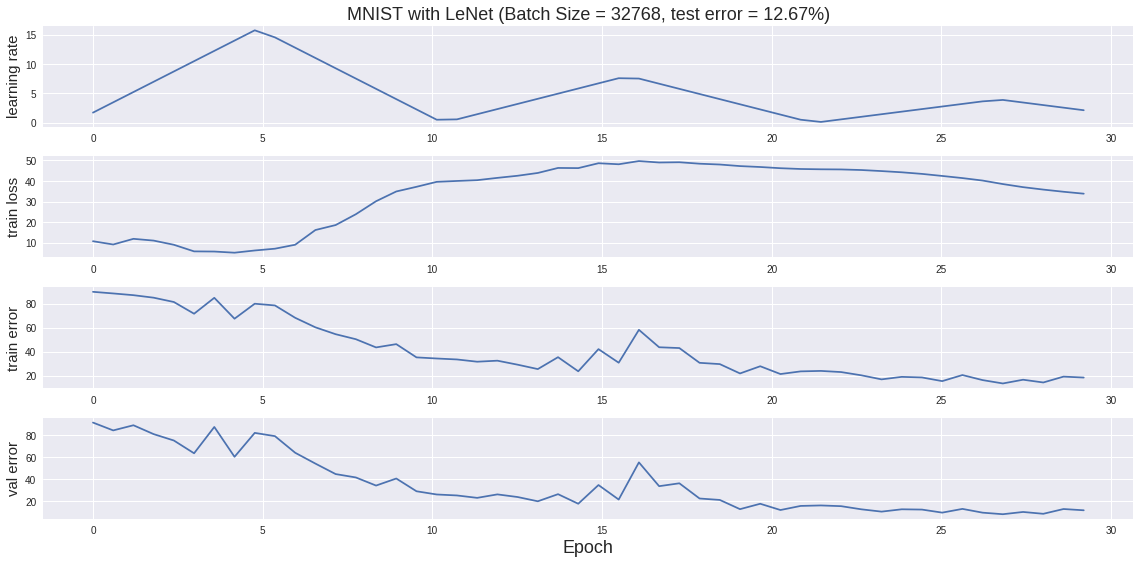}
\caption{LARS optimizer with Cyclical Learning Rate. The baseline with a batch size of 256 can achieve 99.2\% test accuracy (0.8\% test error rate) in 30 epochs.}
\label{fig:lars_mnist_cyc1}
\end{figure}

\begin{figure}
\centering
\includegraphics[width=4.8in]{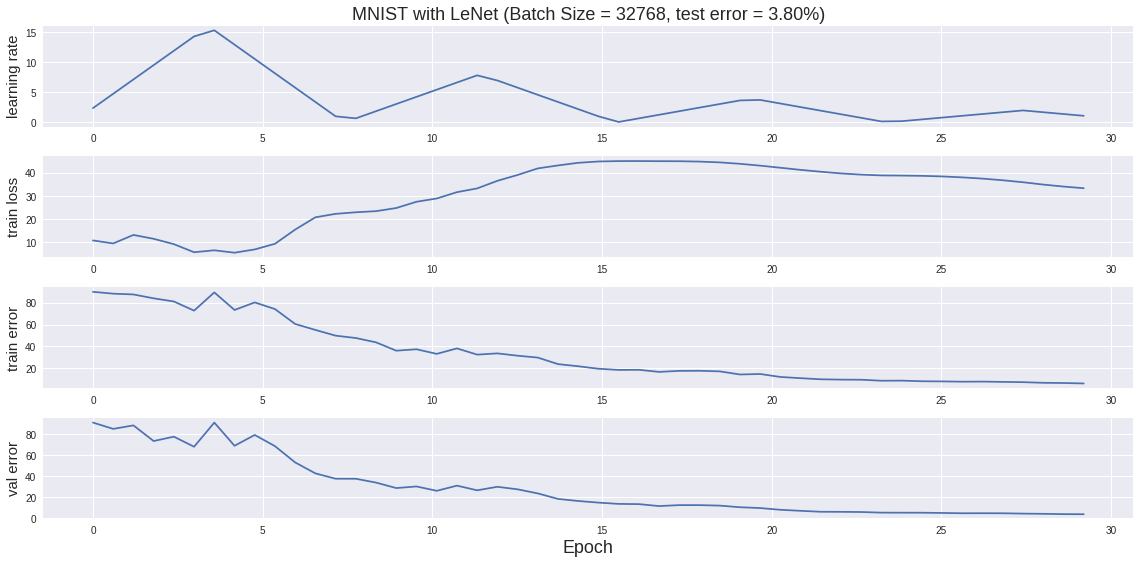}
\caption{LARS optimizer with Cyclical Learning Rate. The baseline with a batch size of 256 can achieve 99.2\% test accuracy (0.8\% test error rate) in 30 epochs. The cyclical length is longer than Figure \ref{fig:lars_mnist_cyc1}.}
\label{fig:lars_mnist_cyc2}
\end{figure}

\subsection{Hardware Usage Discussion}
BERT pre-training is a good example for deep learning applications on supercomputers.
Actually, BERT pre-training can make full use of the most powerful TPU-based supercomputers, which means researchers are able to scale it on 1024 TPU chips \cite{you2019large}.
However, current huge-batch algorithms can not make full use of the supercomputers for some applications like ImageNet. Here, we list a few concrete suggestions for hardware/algorithm designers; what can we do to make better use of supercomputers?

We give the following suggestions:
\begin{itemize}
    \item Using model parallelism and pipeline together with data parallelism, which can maximize the potential performance.
    \item Using a larger sample, which may also be a future trend.
    \begin{itemize}
        \item For ImageNet with ResNet-50, 32 batch size keeps a GPU busy.
        \begin{itemize}
            \item 32K batch size can keep 1024 GPUs busy.
        \end{itemize}
        \item For large image, even a batch size of 1 may keep a GPU busy (e.g. 1920x1080 HD images on ResNet-50).
        \begin{itemize}
            \item 32K batch size can keep 32K GPUs busy.
        \end{itemize}
    \end{itemize}
    \item Trying second-order approach like K-FAC \cite{martens2015optimizing}.
\end{itemize}

\end{document}